\documentclass[10pt,journal,compsoc]{IEEEtran}
\ifCLASSOPTIONcompsoc
 \usepackage[caption=false,font=footnotesize,labelfont=sf,textfont=sf]{subfig}
\else
 \usepackage[caption=false,font=footnotesize]{subfig}
\fi
\usepackage{url}

\ifCLASSOPTIONcompsoc
  \usepackage[nocompress]{cite}
\else
  \usepackage{cite}
\fi
\usepackage{graphicx}
\usepackage{xcolor}
\usepackage{tabu}
\usepackage{bm}
\usepackage{amsmath,amsthm}
\usepackage{amssymb}
\usepackage{tikz}
\usepackage{dsfont}
\usepackage{booktabs}
\usepackage{multirow}
\usepackage{colortbl}
\usepackage[export]{adjustbox}
\usepackage{float}
\usepackage{array}
\usepackage{makecell}
\usepackage{xparse}
\usepackage{mathtools}
\usepackage[linesnumbered,ruled,vlined]{algorithm2e}
\usepackage{cleveref}

\Crefname{algocf}{Algorithm}{Algorithms}

\theoremstyle{plain}
\newtheorem{theorem}{Theorem}

\theoremstyle{plain}
\newtheorem{lemma}[theorem]{Lemma}

\theoremstyle{plain}

\theoremstyle{remark}
\newtheorem{remark}[theorem]{Remark}

\theoremstyle{definition}
\newtheorem{definition}[theorem]{Definition}

\theoremstyle{plain}
\newtheorem{proposition}[theorem]{Proposition}

\DeclarePairedDelimiter\floor{\lfloor}{\rfloor}

\let\emptyset\varnothing

\tikzset{
  treenode/.style = {shape=rectangle, rounded corners,
                     draw, align=center,
                     top color=white,
                     bottom color=blue!20},
  root/.style     = {treenode, font=\Large,
                     bottom color=red!30},
  env/.style      = {treenode, font=\ttfamily\normalsize},
  dummy/.style    = {circle,draw},
  level 1/.style={sibling distance = 16em},
  level 2/.style={sibling distance = 8em},
}

\NewDocumentCommand\Range{O{1}m}{\{ #1,\dots, #2 \}}
\NewDocumentCommand{\todo}{m}{\textcolor{red}{\textbf{Fix:} \emph{#1}\newline}}

\NewDocumentCommand{\change}{}{\textcolor{red}{\textbf{Changes: }}}

\newcommand\kT{k_{\mathcal{T}}}
\newcommand{\td}{\tilde{d}}
\newcommand\norm[1]{\left\lVert#1\right\rVert}
\newcommand{\indep}{\perp \!\!\! \perp}
\newcommand*{\medcap}{\mathbin{\scalebox{1.3}{\ensuremath{\cap}}}}
\newcommand*{\medcup}{\mathbin{\scalebox{1.3}{\ensuremath{\cup}}}}

\DeclareMathOperator*{\argmax}{arg\,max}
\DeclareMathOperator{\diag}{diag}

\newcommand\copyrighttext{%
  \footnotesize \textcopyright 2020 IEEE.  Personal use of this material is permitted. Permission from IEEE must be obtained for all other uses, in any current or future media, including reprinting/republishing this material for advertising or promotional purposes, creating new collective works, for resale or redistribution to servers or lists, or reuse of any copyrighted component of this work in other works.
  DOI: \url{https://dx.doi.org/10.1109/TPAMI.2020.3026019}}
\newcommand\copyrightnotice{%
\begin{tikzpicture}[remember picture,overlay]
\node[anchor=south,yshift=0pt] at (current page.south) {\fbox{\parbox{\dimexpr\textwidth-\fboxsep-\fboxrule\relax}{\copyrighttext}}};
\end{tikzpicture}%
}

\begin{document}
%
\title{Additive Tree-Structured Conditional Parameter Spaces in Bayesian Optimization: A Novel Covariance Function and a Fast Implementation}
%
%
%
%

\author{Xingchen Ma and Matthew B.
 Blaschko
\IEEEcompsocitemizethanks{\IEEEcompsocthanksitem X. Ma and M. B. Blaschko are with the Center for Processing Speech and Images, Department
of Electrical Engineering, KU Leuven, Kasteelpark Arenberg 10, 3001 Leuven, Belgium.\protect\\
E-mail: xingchen.ma@kuleuven.be}
}

%
%

\markboth{Additive Tree-Structured Covariance Function}%
{Shell \MakeLowercase{\textit{et al.}}: Bare Demo of IEEEtran.cls for Computer Society Journals}
%



\IEEEtitleabstractindextext{%
\begin{abstract}
Bayesian optimization (BO) is a sample-efficient global optimization algorithm for black-box functions which are expensive to evaluate. Existing literature on model based optimization in conditional parameter spaces are usually built on trees. In this work, we generalize the additive assumption to tree-structured functions and propose an additive tree-structured covariance function, showing improved sample-efficiency, wider applicability and greater flexibility. Furthermore, by incorporating the structure information of parameter spaces and the additive assumption in the BO loop, we develop a parallel algorithm to optimize the acquisition function and this optimization can be performed in a low dimensional space. We demonstrate our method on an optimization benchmark function, on a neural network compression problem, on pruning pre-trained VGG16 and ResNet50 models as well as on searching activation functions of ResNet20. Experimental results show our approach significantly outperforms the current state of the art for conditional parameter optimization including SMAC, TPE and Jenatton et al. (2017).
\end{abstract}

}

\maketitle
\copyrightnotice

\IEEEdisplaynontitleabstractindextext

%
\IEEEpeerreviewmaketitle



\IEEEraisesectionheading{\section{Introduction}\label{sec:introduction}}
\IEEEPARstart{I}{n} many applications, we are faced with the problem of optimizing an expensive black-box function and we wish to find its optimum using as few evaluations as possible. \textit{Bayesian Optimization} (BO) \cite{jones1998} is a global optimization technique, which is specially suited for these problems. BO has gained increasing attention in recent years \cite{srinivas2010,brochu2010,hutter2011,shahriari2016,frazier2018} and has been successfully applied to sensor location \cite{srinivas2010}, hierarchical reinforcement learning \cite{brochu2010}, and automatic machine learning \cite{klein2017}.

In the general BO setting, we aim to solve the following problem:
\begin{equation*}
    \min_{\bm{x} \in \mathcal{X} \subset \mathbb{R}^d} f(\bm{x}),
\end{equation*}
where $\mathcal{X}$ is a $d$-dimensional parameter space and $f$ is a black-box function which is expensive to evaluate.
Typically, the parameter space $\mathcal{X}$ is treated as structureless, however, for many practical applications, there exists a conditional structure in $\mathcal{X}$: 
\begin{equation}
\label{eq:condition-dep}
    f(\bm{x} \mid \bm{x}_{I_1}) = f(\bm{x}_{I_2} \mid \bm{x}_{I_1}),
\end{equation}
where $I_1,I_2$ are index sets and are subsets of $[d] \coloneqq \{ 1,\dots, d \}$.
\Cref{eq:condition-dep} means given the value of $\bm{x}_{I_1}$, the value of $f(\bm{x})$ remains unchanged after removing $\bm{x}_{[d] \setminus (I_1 \cup I_2)}$. 
Here we use set based subscripts to denote the restriction of $\bm{x}$ to the corresponding indices. 

This paper investigates optimization problems where the parameter space exhibits such a conditional structure. In particular, we focus on a specific instantiation of the general conditional structure in \Cref{eq:condition-dep}: Tree-structured parameter spaces, which are also studied in \cite{jenatton2017}. 
A realistic example is neural network compression. If we compress different layers with different compression algorithms, the search space is a tree-structured parameter space. Another practical example is searching for activation functions of each layer in neural architecture search.

By exploring the properties of this tree structure, we design an \textit{additive tree-structured} (Add-Tree) covariance function, which enables information sharing between different data points under the \textit{additive assumption}, and allows GP to model $f$ in a sample-efficient way. 
Furthermore, by including the tree structure and the additive assumption
in the BO loop, we develop a parallel algorithm to optimize the acquisition function, making the overall execution faster. Our proposed method also helps to alleviate the curse of dimensionality through two advantages: (i) we avoid modeling the response surface directly in a high-dimensional space, and (ii) the acquisition optimization is also operated in a lower-dimensional space.

This article is an extended version of \cite{Ma2020a}.  It has been expanded to include theoretical convergence results of our proposed Add-Tree covariance function in \Cref{sec:theorical-results}, additional practical details in \Cref{sec:practical-consideration}, empirical results of pruning two large-scale neural networks in \Cref{sec:model-pruning} and experiments on neural architecture search in \Cref{sec:resnet20-nas}. 


In the next section, we will briefly review BO together with the literature related to optimization in a conditional parameter space. 
In Section~\ref{sec:problem_formulation}, we formalize the family of objective functions that can be solved using our approach. We then present our Add-Tree covariance function in Section~\ref{sec:add-tree-kernel}. 
In Section~\ref{sec:bo}, we give the inference procedure and BO algorithm using our covariance function. We further present theoretical properties of an Add-Tree covariance function in \Cref{sec:theorical-results}. We then report a range of experiments in Section~\ref{sec:experiment}. Finally, we conclude in Section~\ref{sec:conclusion}. All proofs and implementation details\footnote{Code available at \url{https://github.com/maxc01/addtree}} are provided in the supplementary material.

\subsection{Related Work}
\label{sec:related_work}

\subsubsection{Bayesian Optimization}
\label{subsec:bo}

BO has two major components. The first one is a probabilistic regression model used to fit the response surface of $f$. Popular choices include GPs \cite{brochu2010}, random forests \cite{hutter2011} and adaptive Parzen estimators \cite{bergstra2011}. We refer the reader to \cite{rasmussen2006} for the foundations of Gaussian Processes. The second one is an acquisition function $u_{t}$ which is constructed from this regression model and is used to propose the next evaluation point. Popular acquisition functions include the \textit{expected improvement} (EI) \cite{jones1998}, \textit{knowledge gradient} (KG) \cite{frazier2009}, \textit{entropy search} (ES) \cite{hennig2012} and \textit{Gaussian process upper confidence bound} (GP-UCB) \cite{srinivas2010}. 

One issue that often occurs in BO is, in high-dimensional parameter spaces, its performance may be no better than random search \cite{wang2013,li2016}. This deterioration is because of the high uncertainty in fitting a regression model due to the curse of dimensionality \cite[Ch.~2]{gyorfi2002}, which in turn leads to pure-explorational behavior of BO. 
This will further cause inefficiency in the acquisition function, making the proposal of the next data point behave like random selection. 
Standard GP-based BO ignores the structure in a parameter space, and fits a regression model in $\mathbb{R}^d$. By leveraging this structure information, we can work in a low-dimensional space $\mathbb{R}^m$ (recall \Cref{eq:condition-dep}) instead of $\mathbb{R}^d$. 
Another well-known issue of high-dimensional BO is the problem of optimizing an acquisition function is also performed in a high-dimensional space. This difficulty can also be sidestepped by using this structure information if GP-UCB is used as our acquisition function.



\subsubsection{Conditional Parameter Spaces}
\label{subsec:cps}

Sequential Model-based Algorithm Conﬁguration (SMAC)~\cite{hutter2011} and Tree-structured Parzen Estimator Approach (TPE)~\cite{bergstra2011} are two popular non-GP based optimization algorithms that are aware of the conditional structure in $\mathcal{X}$, however, they lack favorable properties of GPs: uncertainty estimation in SMAC is non-trivial and the dependencies between dimensions are ignored in TPE. Additionally, neither of these methods have a particular sharing mechanism, which is valuable in the low-data regime. 

In the category of GP-based BO, which is our focus in this paper, \cite{hutter2013} proposed a covariance function that can explicitly employ the tree structure and share information at those categorical nodes. However, their specification for the parameter space is too restrictive and they require the shared node to be a categorical variable.  By contrast, we allow shared variables to be continuous (see Section~\ref{sec:add-tree-kernel}). \cite{swersky2014a} applied the idea of \cite{hutter2013} in a BO setting, but their method still inherits the limitations of \cite{hutter2013}.  Another covariance function to handle tree-structured dependencies is presented in \cite{levesque2017}. In that case, they force the similarity of two samples from different condition branches to be zero and the resulting model can be transformed into several independent GPs.  We perform a comparison to an independent GP baseline in Section~\ref{sec:synthetic-function}. In contrast to Add-Tree, the above approaches either have very limited applications, or lack a sharing mechanism. 
\cite{jenatton2017} presented another GP-based BO approach, where they handle  tree-structured dependencies by introducing a weight vector linking all sub-GPs, and this introduces an explicit sharing mechanism. Although \cite{jenatton2017} overcame the above limitations, the enforced linear relationships between different paths make their semi-parametric approach less flexible compared with our method. We observe in our experiments that this can lead to a substantial difference in performance.

\section{Problem Formulation}
\label{sec:problem_formulation}

\begin{figure}[t]
\centering
\begin{tikzpicture}[
    grow                    = right,
    edge from parent/.style = {draw, -latex},
    root/.style     = {treenode, font=\large, bottom color=red!30},
    every node/.style       = {font=\footnotesize},
    level distance          = 10em,
    level 1/.style={sibling distance = 8em},
    sloped
    ]
  \node [root] {$r, \bm{v}_r \in [-1,1]^2$}
    child { node [env] {$p_1, \bm{v}_{p_1} \in [-1,1]^2, f_{p_1, \mathcal{T}} = \norm{\bm{v}_r}^2 + \norm{\bm{v}_{p_1}}^2 $} edge from parent node[above] {t=1}  }
    child { node [env] {$p_2, \bm{v}_{p_2} \in [-1,1]^3, f_{p_2, \mathcal{T}} = \norm{\bm{v}_r}^2 + \norm{\bm{v}_{p_2}}^2 $} edge from parent node[below] {t=2}  };
\end{tikzpicture}
\caption{A simple tree-structured function. The dimension of $\mathcal{X}$ is $d=8$, and the effective dimension at $p_1$ and $p_2$ are 4 and 5 respectively.} 
\label{fig:example}
\end{figure}
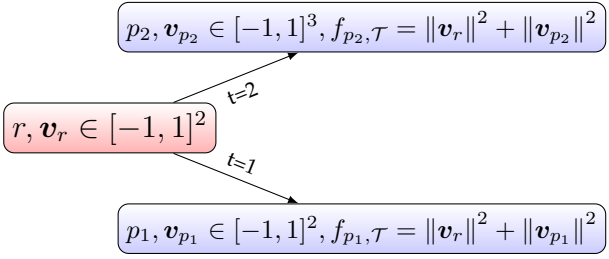

We begin by listing notation used in this paper. Vectors, matrices and sets will be denoted by bold lowercase letters, bold uppercase letters and regular uppercase letters respectively. Let $\mathcal{T}=(V,E)$ be a tree, in which $V$ is the set of vertices, $E$ is the set of edges, $P=\{p_i\}_{1\leq i \leq |P|}$ be the set of leaves and $r$ be the root of $\mathcal{T}$ respectively, $\{ l_i \}_{1\leq i \leq |P|}$ be the ordered set of vertices on the path from $r$ to the $i$-th leaf $p_{i}$, and $h_i$  be the number of vertices along $l_i$ (including $r$ and $p_i$). To distinguish an objective function defined on a tree-structured parameter space from a general objective function, we use $f_{\mathcal{T}}$ to indicate our objective function. In what follows, we will call $f_{\mathcal{T}}$ a~\textit{tree-structured function}. An example of such a tree-structured function is depicted in \Cref{fig:example}, where each node of $\mathcal{T}$ is associated with a bounded continuous variable and each edge represents a specific setting of a categorical variable.



\begin{definition}[Tree-structured parameter space]
\label{def:TreeStructuredParameterSpace}
A \emph{tree-structured parameter space} $\mathcal{X}$ is associated with a tree $\mathcal{T}=(V,E)$. For any $v \in V$, $v$ is associated with a bounded continuous variable of $\mathcal{X}$; for any $e \in E$, 
the set of outgoing edges $E_v$ of $v$ represent one categorical variable of $\mathcal{X}$ and each element of $E_v$ represents a specific setting of the corresponding categorical variable.
\end{definition}

\begin{definition}[Tree-structured function]
\label{def:TreeStructuredFunction}
A \emph{tree-structured function} $f_{\mathcal{T}}: \mathcal{X} \rightarrow \mathbb{R}$ is defined on a $d$-dimensional tree-structured parameter space $\mathcal{X}$. The $i$-th leaf $p_i$ is associated with a function $f_{p_i,\mathcal{T}}$ of the variables associated with the vertices along $l_i$. $f_{\mathcal{T}}$ is called \emph{tree-structured} if for every leaf of the tree-structured parameter space
\begin{equation}
\label{eq:def-tree-structured-function}
f_{\mathcal{T}}(\bm{x}) \coloneqq f_{p_j,\mathcal{T}}(\bm{x}\vert_{l_j}) ,
\end{equation}
where $p_j$ is selected by the categorical values of $\bm{x}$ and $\bm{x}\vert_{l_j}$ is the restriction of $\bm{x}$ to $l_j$.
\end{definition}


To aid in the understanding of a tree-structured function (and subsequently our proposed Add-Tree covariance function), we depict a simple tree-structured function in \Cref{fig:example}. 
The categorical variable in this example is $t\in \{1,2\}$ and its values are shown around the outgoing edges of $r$. The continuous variables are $\bm{v}_r \in [-1,1]^2, \bm{v}_{p_1} \in [-1,1]^2, \bm{v}_{p_2} \in [-1,1]^3$ and they are displayed inside vertices $r,p_1,p_2$ respectively. Here $\bm{v}_r$ is a shared continuous parameter, as evidenced by it appearing in both leaves. Two functions $f_{p_1,\mathcal{T}}$ and $f_{p_2,\mathcal{T}}$ associated with leaves $p_1$ and $p_2$ respectively are also shown in \Cref{fig:example}.
In \Cref{def:TreeStructuredFunction}, the restriction of one input to a path means we collect variables associated with the vertices along that path and concatenate them using a fixed ordering.  
For example, in \Cref{fig:example}, let $\bm{x} \in \mathcal{X}$ be an 8-dimensional input, then the restriction of $\bm{x}$ to path $l_1$ is a 4-dimensional vector. The function illustrated in \Cref{fig:example} can be compactly written down as:
\begin{align} \label{eq:FigureExampleTreeFunction}
    f_{\mathcal{T}}(\bm{x}) = \mathds{1}_{t=1} f_{p_1,\mathcal{T}}(\bm{x}\vert_{l_1}) + \mathds{1}_{t=2} f_{p_2,\mathcal{T}}(\bm{x}\vert_{l_2}) , 
\end{align}
where $\bm{x}$ is the concatenation of $(\bm{v}_r,\bm{v}_{p_1},\bm{v}_{p_2},t)$, $\mathds{1}$ denotes the indicator function, $f_{p_1,\mathcal{T}}(\bm{x}\vert_{l_1})=\norm{\bm{v}_r}^2 + \norm{\bm{v}_{p_1}}^2$ and $f_{p_2,\mathcal{T}}(\bm{x}\vert_{l_2}) = \norm{\bm{v}_r}^2 + \norm{\bm{v}_{p_2}}^2$.

A tree-structured function $f_{\mathcal{T}}$ is actually composed of several smaller functions $\{ f_{p_i,\mathcal{T}} \}_{1 \leq i \leq |P|}$, given a specific setting of the categorical variables, $f_{\mathcal{T}}$ will return the associated function at the $i$-th leaf. To facilitate our description in the following text, we define the \textit{effective dimension} in \Cref{def:EffectiveDimension}.

\begin{definition}[Effective dimension]
\label{def:EffectiveDimension}
The \emph{effective dimension} of a tree-structured function $f_{\mathcal{T}}$ at the $i$-th leaf $p_i$ is the sum of dimensions of the variables associated with the vertices along $l_i$. 
\end{definition}

\begin{remark}
The effective dimension of $f_{\mathcal{T}}$ can be much smaller than the dimension of $\mathcal{X}$. 
\end{remark}

Particularly, if $\mathcal{T}$ is a perfect binary tree, in which each vertex is associated with a 1-dimensional continuous variable, the effective dimension of $f_{\mathcal{T}}$ at every leaf is the depth $h$ of $\mathcal{T}$, while the dimension of $\mathcal{X}$ is $3 \cdot 2^{h-1}-2$.
If the tree structure information is thrown away, we have to work in a much higher dimensional parameter space. 
It is known that in high dimensions, BO behaves like random search~\cite{wang2013,li2016}, which violates the entire purpose of model based optimization. 

Now we have associated the parameter space $\mathcal{X}$ with a tree structure, which enables us to work in the low-dimensional effective space. How can we leverage this tree structure to optimize $f_{\mathcal{T}}$? Recalling $f_{\mathcal{T}}$ is a collection of $|P|$ functions $\{ f_{p_i,\mathcal{T}} \}_{1 \leq i \leq |P|}$, a trivial solution is using $|P|$ independent GPs to model each $f_{p_i, \mathcal{T}}$ separately. 
In BO settings, we are almost always in low data regime because black-box calls to $f_{\mathcal{T}}$ are expensive (e.g.\ the cost of training and evaluating a machine learning model). Modeling $f_{\mathcal{T}}$ using a collection of GPs is obviously not an optimal way because we discard the correlation between $f_{p_i,\mathcal{T}}$ and $f_{p_j,\mathcal{T}}$ when $i \neq j$. How to make the most of the observed data, especially how to share information between data points coming from different leaves remains a crucial question. In this paper, we assume additive structure within each $f_{p_i,\mathcal{T}}$ for $i=1,\cdots,|P|$. More formally, $f_{p_i,\mathcal{T}}$ can be decomposed in the following form:
\begin{equation}
\label{eq:additive-assumption}
    f_{p_i,\mathcal{T}}(\bm{x}) = \sum_{j=1}^{h_i} f_{ij}(v_{ij})
\end{equation}
where $v_{ij}$ is the associated variable on the $j$-th vertex along $l_i$.
The additive assumption has been extensively studied in the GP literature \cite{duvenaud2011,kandasamy2015,gardner2017,rolland2018} and is a popular method of dimension reduction \cite{gyorfi2002}.
We note the tree-structured function discussed in this paper is a generalization of the objective function presented in these publications and our additive assumption in \Cref{eq:additive-assumption} is also a generalization of the additive structure considered previously. For example, the additive function discussed in \cite{kandasamy2015} can be viewed as a tree-structured function the associated tree of which has a branching factor of 1, i.e.\ $|P|=1$.
Our generalized additive assumption will enable an efficient sharing mechanism, as we develop in Section~\ref{sec:add-tree-kernel}.

\section{THE ADD-TREE COVARIANCE FUNCTION}
\label{sec:add-tree-kernel}

In this section, we describe how we use the tree structure and the additive assumption to design a covariance function, which is sample-efficient in low-data regime. We start with the definition of the Add-Tree covariance function (Definition~\ref{def:Add-Tree}), then we show the intuition (\Cref{eq:joint-distributation}) behind this definition and present an algorithm (Algorithm~\ref{algo:kernel-construction}) to automatically construct an Add-Tree covariance function from the specification of a tree-structured parameter space, finally a proof of the validity of this covariance function is given. 


\begin{definition}[Add-Tree covariance function]
\label{def:Add-Tree}
For a tree-structured function $f_{\mathcal{T}}$, let $\bm{x}_{i'}$ and $\bm{x}_{j'}$  be two inputs of $f_{\mathcal{T}}$, $p_i$ and $p_j$ be the corresponding leaves, $a_{ij}$ be the lowest common ancestor (LCA) of $p_i$ and $p_j$, $l_{ij}$ be the path from $r$ to $a_{ij}$. A covariance function $k_{f_{\mathcal{T}}}: \mathcal{X} \times \mathcal{X} \rightarrow \mathbb R $ is said to be an \emph{Add-Tree covariance function} if for each $\bm{x}_{i'}$ and $\bm{x}_{j'}$
\begin{align}
\label{eq:add-tree-kernel-1}
    k_{\mathcal{T}}(\bm{x}_{i'},\bm{x}_{j'}) \coloneqq & k_{l_{ij}}(\bm{x}_{i'}\vert_{l_{ij}},\bm{x}_{j'}\vert_{l_{ij}})  \nonumber \\
    = & \sum_{v \in l_{ij}} k_v(\bm{x}_{i'}\vert_{v},\bm{x}_{j'}\vert_{v})
\end{align}
where $\bm{x}_{i'}\vert_{l_{ij}}$ is the restriction of $\bm{x}_{i'}$ to the variables along the path $l_{ij}$, and 
$k_v$ is any positive semi-definite covariance function on the continuous variables appearing at a vertex $v$ on the path $l_{ij}$. 
We note the notation $l_{ij}$ introduced here is different from the notation $l_i$ introduced in the beginning of \Cref{sec:problem_formulation}.
\end{definition}


We take the tree-structured function illustrated in \Cref{fig:example} (\Cref{eq:FigureExampleTreeFunction}) as an example.\footnote{To simplify the presentation, we use a two-level tree structure in this example. The covariance function, however, generalizes to tree-structured functions of arbitrary depth (Algorithm~\ref{algo:kernel-construction}).} 
Suppose there are $n_1$ and $n_2$ data points in $l_1$ and $l_2$ respectively, and let $X_1 \in \mathbb{R}^{n_1 \times d_1} $ and $X_2 \in \mathbb{R}^{n_2 \times d_2}$ be the inputs from $l_1$ and $l_2$, where $d_1=2+2$ and $d_2=2+3$ are the effective dimensions of $f_{\mathcal{T}}$ at $p_1$ and $p_2$ respectively. Denote the latent variables associated to the decomposed functions\footnote{On functions and latent variables, one can refer to~\cite[chap.~2]{rasmussen2006}.} at $r$, $p_1$ and $p_2$ by $\bm{f}_r \in \mathbb{R}^{n_1+n_2}$, $\bm{f}_1 \in \mathbb{R}^{n_1}$ and $\bm{f}_2 \in \mathbb{R}^{n_2}$, respectively. Reordering and partition $\bm{f}_r$ into two parts corresponding to $p_1$ and $p_2$, so that 
\begin{equation*}
    \bm{f}_{r} = 
    \begin{bmatrix}
           \bm{f}_{r}^{(1)} \\
           \bm{f}_{r}^{(2)}
    \end{bmatrix}, 
    \bm{f}_{r}^{(1)} \in \mathbb{R}^{n_1}, 
    \bm{f}_{r}^{(2)} \in \mathbb{R}^{n_2}.
\end{equation*}
Let the Gram matrix corresponding to $\bm{f}_r, \bm{f}_1, \bm{f}_2$ be $\bm{K}_r \in \mathbb{R}^{(n_1+n_2) \times (n_1+n_2)}, \bm{K}_1 \in \mathbb{R}^{n_1 \times n_1}, \bm{K}_2 \in \mathbb{R}^{n_2 \times n_2}$. W.l.o.g, let the means of $\bm{f}_r,\bm{f}_1,\bm{f}_2$ be $\bm{0}$. By the additive assumption in \Cref{eq:additive-assumption}, the latent variables corresponding to the associated functions at $p_1$ and $p_2$ are $\bm{f}_{r}^{(1)} + \bm{f}_1$ and $\bm{f}_{r}^{(2)} + \bm{f}_2$, we have:
\begin{equation}
\label{eq:step1}
    \begin{bmatrix}
           \bm{f}_{r}^{(1)} + \bm{f}_1 \\
           \bm{f}_{r}^{(2)} + \bm{f}_2 
         \end{bmatrix} = \begin{bmatrix}
           \bm{f}_{r}^{(1)} \\
           \bm{f}_{r}^{(2)}
         \end{bmatrix} + \begin{bmatrix}
           \bm{f}_1 \\
           \bm{f}_2 
         \end{bmatrix}, 
             \begin{bmatrix}
           \bm{f}_{r}^{(1)} \\
           \bm{f}_{r}^{(2)}
         \end{bmatrix} \sim \mathcal{N}(\bm{0}, \bm{K}_r) .
\end{equation}
Due to $\bm{f}_1 \indep \bm{f}_2$, where $\indep$ denotes $\bm{f}_1$ is independent of $\bm{f}_2$, we have:
\begin{equation}
\label{eq:step2}
    \begin{bmatrix}
           \bm{f}_1 \\
           \bm{f}_2 \\
         \end{bmatrix} \sim \mathcal{N} \left( \bm{0}, 
         \begin{bmatrix}
           \bm{K}_1  & \bm{0}  \\
            \bm{0}    & \bm{K}_2 
         \end{bmatrix} \right), 
\end{equation}
furthermore, because of the additive assumption in \Cref{eq:additive-assumption},
\begin{equation}
\label{eq:step3}
    \begin{bmatrix}
           \bm{f}_{r}^{(1)} \\
           \bm{f}_{r}^{(2)}
         \end{bmatrix} \indep 
         \begin{bmatrix}
           \bm{f}_1 \\
           \bm{f}_2 
         \end{bmatrix} .
\end{equation}
Combining \Cref{eq:step1,eq:step2,eq:step3}, we arrive at our key conclusion:
\begin{equation}
        \begin{bmatrix}
           \bm{f}_r^{(1)} {+} \bm{f}_1 \\
           \bm{f}_r^{(2)} {+} \bm{f}_2 \\
         \end{bmatrix} \sim \mathcal{N}\left(\bm{0}, 
         \left[
         \begingroup 
         \begin{array}{ll}
           \bm{K}_r^{(11)} + \bm{K}_1 & \bm{K}_r^{(12)} \\
           \bm{K}_r^{(21)}       & \bm{K}_r^{(22)} + \bm{K}_2
         \end{array}
         \endgroup
         \right]
         \right),
\label{eq:joint-distributation}         
\end{equation}
where $\bm{K}_r$ is decomposed as follows:
\begin{equation*}
    \bm{K}_r = \begin{bmatrix} 
    \bm{K}_r^{(11)} & \bm{K}_r^{(12)} \\ 
    \bm{K}_r^{(21)} & \bm{K}_r^{(22)} 
    \end{bmatrix}
\end{equation*}
in which $\bm{K}_r^{(11)} \in \mathbb{R}^{n_1 \times n_1}, \bm{K}_r^{(12)} \in \mathbb{R}^{n_1 \times n_2}, \bm{K}_r^{(21)} \in \mathbb{R}^{n_2 \times n_1}, \bm{K}_r^{(22)} \in \mathbb{R}^{n_2 \times n_2}$. The observation in \Cref{eq:joint-distributation} is crucial in two aspects: firstly, we can use a single covariance function and a global GP to model our objective, secondly and more importantly, this covariance function allows an efficient sharing mechanism between data points coming from different paths, although we cannot observe the decomposed function values at the shared vertex $r$, we can directly read out this sharing information from $\bm{K}_r^{(12)}$.

\begin{algorithm}
\DontPrintSemicolon

\SetKwFunction{Kernel}{Kernel}
\SetKwFunction{Index}{Index}
\SetKwFunction{Dim}{Dim}
\SetKwFunction{Range}{Range}
\SetKwInOut{Input}{Input}
\SetKwInOut{Output}{Output}

\Input{The associated tree $\mathcal{T}=(V,E)$ of $f_\mathcal{T}$}
\Output{Add-Tree covariance function $k_{\mathcal{T}}$}


$k_{\mathcal{T}} \gets 0$ \;
\For{$v \gets V$}{
  $vi \gets \Index{v} $ \;
  $k_v^{d} \gets k_{v}^{\delta}(\bm{x}, \bm{x}') $ \tcc*{$k_{v}^{\delta}(\bm{x},\bm{x}')=1$ iff $\bm{x}$ and $\bm{x}'$ both contain $v$ in their paths}
  $si \gets vi + 1$ \tcc*{start index of v}
  $ei \gets vi + 1 + \Dim{v}$ \tcc*{end index of v}
  $k_v^{c} \gets k_c(\bm{x}_{si \leq i < ei}, \bm{x}'_{si \leq i < ei})$ \tcc*{$k_c$ is any p.s.d.\ covariance function}
  $k_v \gets k_v^{d} \times k_{v}^{c}$ \;
  $k_{\mathcal{T}} \gets k_{\mathcal{T}} + k_v$ 
}
\caption{Add-Tree Covariance Function}
\label{algo:kernel-construction}
\end{algorithm}

We summarize the  construction of an  Add-Tree covariance function in \Cref{algo:kernel-construction}. We note \Cref{algo:kernel-construction} is used to construct a function object of an Add-Tree covariance function, and during this construction stage, no observations are involved in the computation.
The $7$-th line in \Cref{algo:kernel-construction} needs further explanation. If $\bm{x}$ and $\bm{x}'$ don't intersect, at least one of $\bm{x}_{si \leq i < ei}$ and $\bm{x}'_{si \leq i < ei}$ is empty (it is possible that both of them are empty). In this case, we let $k_c$ map an empty set to any non-negative scalar, and a common choice is 0. In practice, when $\bm{x}$ and $\bm{x}'$ don't contain vertex $v$ at the same time, the value of $k_c(\bm{x}_{si \leq i < ei}, \bm{x}'_{si \leq i < ei})$ is irrelevant because it is not needed. 


\begin{proposition}
\label{prop:add-tree-is-psd}
The Add-Tree covariance function defined by Definition~\ref{def:Add-Tree} is positive semi-definite for all tree-structured functions defined in Definition~\ref{def:TreeStructuredFunction} with the additive assumption satisfied. 
\end{proposition}

\section{BO for Tree-Structured Functions}
\label{sec:bo}

In this section, we first describe how to perform the inference with our proposed Add-Tree covariance function, then we present a parallel algorithm to optimize the acquisition function. At the end of this section, we give the convergence rate of an Add-Tree covariance function. 

\subsection{Inference with Add-Tree}

Given noisy observations $D_{n} = \{ (\bm{x}_{i}, y_{i}) \}_{i=1}^{n}$, we would like to obtain the posterior distribution of $f_{\mathcal{T}}(\bm{x}_*)$ at a new input $\bm{x}_{*} \in \mathcal{X}$.
We begin with some notation. Let $p_{*}$ be the leaf selected by the categorical values of $\bm{x}_{*}$, $l_{*}$ be the path from the root $r$ to $p_{*}$. All $n$ inputs are collected in the design matrix 
$\bm{X} \in \mathbb{R}^{n \times d}$, where the $i$-th row represents $\bm{x}_i \in \mathbb{R}^d$, and the targets and observation noise are aggregated in vectors $\bm{y} \in \mathbb{R}^n$ and $\bm{\sigma} \in \mathbb{R}^n$ respectively. 
Let $\bm{\Sigma}=\diag(\bm{\sigma})$ be the noise matrix, where $\diag(\bm{\sigma})$ denotes a diagonal matrix containing the elements of vector $\bm{\sigma}$, 
$S=\{i \mid l_{*} \medcap l_{i} \neq \emptyset \}$, 
$\bm{I} \in \mathbb{R}^{n \times n}$ be the identity matrix, $\bm{Q} \in \mathbb{R}^{|S| \times n}$ be a selection matrix, which is constructed by removing the $j$-th row of $\bm{I}$ if $j \notin S$, $\bm{X}' = \bm{Q} \bm{X} \in \mathbb{R}^{|S| \times d}$, $\bm{y}' = \bm{Q} \bm{y} \in \mathbb{R}^{|S|}$, $\bm{\Sigma}^{'} = \bm{Q} \bm{\Sigma} \bm{Q}^T \in \mathbb{R}^{|S| \times |S|}$. 
We note it is possible that a vertex is not associated with any continuous variable. In particular, if there is no continuous variable associated with root $r$,  $|S|$ is smaller than $n$. 
Then we can write down the joint distribution of $f_{\mathcal{T}}(\bm{x}_*)$ and $\bm{y}'$ as:
\begin{equation*}
    \begin{bmatrix}
           f_{\mathcal{T}}(\bm{x}_*) \\
           \bm{y'}
    \end{bmatrix} \sim \mathcal{N}\left(\bm{0},
        \begin{bmatrix}
        k_{\mathcal{T}}(\bm{x}_{*}, \bm{x}_{*}) & \bm{K}_{\mathcal{T}}(\bm{x}_{*}) \\
        \bm{K}_{\mathcal{T}}(\bm{x}_{*})^T & \bm{K}_{\mathcal{T}} + \bm{\Sigma}'
        \end{bmatrix}
    \right),
\end{equation*}
where $\bm{K}_{\mathcal{T}} \coloneqq \bm{K}_{\mathcal{T}}(\bm{X}') = [k_{\mathcal{T}}( \bm{x}, \bm{x}')]_{\bm{x}, \bm{x}' \in \bm{X}'}$, $\bm{K}_{\mathcal{T}}(\bm{x}_{*}) = [k_{\mathcal{T}}(\bm{x}_{*}, \bm{x})]_{\bm{x} \in \bm{X}'}$. 
We note that this joint distribution has the same standard form~\cite{rasmussen2006} as in all GP-based BO, but that it is made more efficient by the selection of $\bm{X}'$ based on the tree structure. The predictive distribution of $f_{\mathcal{T}}(\bm{x}_*)$ is: 
\begin{equation}
    f_{*\mathcal{T}} \mid \bm{X}',\bm{y}',\bm{x}_{*} \sim \mathcal{N}(\mu_{\mathcal{T}}(\bm{x}_*), \sigma_{\mathcal{T}}^2(\bm{x}_*))
\label{eq:predictive-distribution}
\end{equation}
where
\begin{eqnarray*}
\mu_{\mathcal{T}}(\bm{x}_*) & = &  \bm{K}_{\mathcal{T}}(\bm{x}_{*}) \bm{K}_{\bm{y},\mathcal{T}}^{-1} \bm{y}, \\
\sigma_{\mathcal{T}}^2(\bm{x}_*) & = &  k_{\mathcal{T}}(\bm{x}_{*}, \bm{x}_{*}) - \bm{K}_{\mathcal{T}}(\bm{x}_{*}) \bm{K}_{\bm{y},\mathcal{T}}^{-1} \bm{K}_{\mathcal{T}}(\bm{x}_{*})^T, \\
\bm{K}_{\bm{y},\mathcal{T}} & = & \bm{K}_{\mathcal{T}} + \bm{\Sigma}' .
\end{eqnarray*}


Black-box calls to the objective function usually dominate the running time of BO, and the time complexity of fitting GP is of less importance in BO settings. Details on the time complexity analysis of an Add-Tree covariance function can be found in \cite{Ma2020a}, and we only sketch the idea here. The time complexity of inference using an Add-Tree covariance function can be derived by recursion. W.l.o.g, let $\mathcal{T}$ be a binary tree. If root $r$ of $\mathcal{T}$ is associated with a continuous parameter, in this worst case, the complexity is $\mathcal{O}(n^3)$, where $n$ is the number of observations. Otherwise, if $r$ is not associated to any continuous parameter, let $n_l$ and $n_r$ be the number of samples falling into the left and the right path respectively, we have $T(r) = T(r_l) + T(r_r)$, where $r_l$ and $r_r$ are the left and the right child of $r$ respectively, and $T(r),T(r_l),T(r_r)$ are the running time at nodes $r,r_l,r_r$ respectively. Because the worst-case running time at nodes $r_l$ and $r_r$ is $\mathcal{O}(n_l^3)$ and $\mathcal{O}(n_r^3)$ respectively, we have $T(r)=\mathcal{O}(n_l^3 + n_r^3)$.

\subsection{Acquisition Function Optimization}                 

In BO, the acquisition function $u_{t-1}(\bm{x}| D_{t-1})$, where $t$ is the current step of optimization, is used to guide the search for the optimum of an objective function. By trading off the exploitation and exploration, it is expected we can find the optimum using as few calls as possible to the objective. 
To get the next point at which we evaluate an objective function, we solve $\bm{x}_{t} = \argmax_{\bm{x} \in \mathcal{X}} u_{t-1}(\bm{x}| D_{t-1})$. 
For noisy observations, GP-UCB~\cite{srinivas2010} is proved to be no-regret and has explicit regret bounds for many commonly used covariance functions, and in this work, we will use GP-UCB, which is defined as:
\begin{equation*}
    u_{t-1}(\bm{x}| D_{t-1}) =  \mu_{t-1}(\bm{x}) + \beta_{t-1}^{1/2} \sigma_{t-1}(\bm{x}),
\end{equation*}
where $\beta_{t-1}$ are suitable constants which will be given later in \Cref{sec:theorical-results}, $\mu_{t-1}(\bm{x})$ and $\sigma_{t-1}(\bm{x})$ are the predictive posterior mean and standard deviation at $\bm{x}$ from \Cref{eq:predictive-distribution}. 


A na\"{i}ve way to obtain the next evaluation point for a tree-structured function is to independently find $|P|$ optima, each one corresponding to the optimum of the associated function at a leaf, and then choose the best candidate across these optima. This approach is presented in~\cite{jenatton2017} and the authors there already pointed out this is too costly. Here we develop a much more efficient algorithm, which is dubbed as Add-Tree-GP-UCB, to find the next point and we summarise it in Algorithm~\ref{algo:bo}. By explicitly utilizing the associated tree structure $\mathcal{T}$ of $\bm{f}_{\mathcal{T}}$ and the additive assumption, the first two nested \texttt{for} loops can be performed in parallel.  Furthermore, as a by-product, each acquisition function optimization routine is now performed in a low dimensional space whose dimension is even smaller than the effective dimension. 
We note the Add-Tree covariance function developed here can be combined with any other acquisition function, however, \Cref{algo:bo} may not be applicable. 


\begin{algorithm}
\DontPrintSemicolon

\SetKwFunction{Kernel}{Kernel}
\SetKwFunction{Index}{Index}
\SetKwFunction{Dim}{Dim}
\SetKwFunction{Range}{Range}
\SetKwFunction{Pop}{Pop}
\SetKwFunction{Concat}{Concat}

\SetKwInOut{Input}{Input}
\SetKwInOut{Output}{Output}
\Input{The associated tree $\mathcal{T}=(V,E)$ of $f_\mathcal{T}$, Add-Tree covariance function $k_{\mathcal{T}}$ from Algorithm \ref{algo:kernel-construction}, paths $\{ l_i \}_{1\leq i \leq |P|}$ of $\mathcal{T}$}

$D_0 \gets \emptyset$\;
\For{$t \gets 1,\dots$}{
  \For{$v \gets V$}{
    $\bm{x}_{t}^{v} \gets \argmax_x  \mu_{t-1}^{v}(\bm{x}) + \sqrt{\beta_t} \sigma_{t-1}^{v}(\bm{x})$ \;
    $u_{t}^{v} \gets \mu_{t-1}^{v}(\bm{x}_{t}^{v}) + \sqrt{\beta_t} \sigma_{t-1}^{v}(\bm{x}_{t}^{v})$ \;
  }
  
  \For{$i \gets 1,\dots,|P|$}{
    $U_t^{l_i} \gets \sum_{v \in l_i} u_{t}^{v} $ \tcc*{additive assumption from \Cref{eq:additive-assumption}}
  }
  $j \gets \argmax_{i} \{ U_t^{l_i} \mid i=1,\dots,|P|\}  $ \;
  $\bm{x}_t \gets \medcup \{\bm{x}_t^{v} \vert v \in l_j\}$ \;
  $y_{t} \gets \bm{f}_{\mathcal{T}}(\bm{x}_t)$ \;
  $D_t \gets D_{t-1} \medcup \{(\bm{x}_t,y_t) \}$ \;
  Fitting GP using $D_t$ to get $\{ (\mu_t^{v},\sigma_t^{v}) \}_{v \in V}$ \;
}

\caption{Add-Tree-GP-UCB}
\label{algo:bo}
\end{algorithm}

\subsection{Theoretical Results}
\label{sec:theorical-results}

In this section, we give theoretical results regarding our proposed Add-Tree covariance function. Let $r_t = f_{\mathcal{T}}(\bm{x}^{*}) - f_{\mathcal{T}}(\bm{x}_t)$ be the instantaneous regret, where $f_{\mathcal{T}}(\bm{x}^{*}) = \max_{\bm{x} \in \mathcal{X}} f_{\mathcal{T}}(\bm{x})$, and let $R_t = \sum_{i=1}^{t} r_i$ be the cumulative regret after $t$ iterations \cite{cesa-bianchi2006}. 
Similar to \cite{berkenkamp2019}, we show our proposed Add-Tree covariance function has sublinear regret bound if each covariance function defined on the vertices of $\mathcal{T}$ is a squared exponential kernel or a Mat\'{e}rn kernel. 
We further show without taking the structure of the parameter space into consideration, the cumulative regret has an exponential dependence with the dimension $d$ of $\mathcal{X}$, while our method has an exponential dependence with the highest dimension $\td$ of functions defined in vertices of $\mathcal{T}$. As stated earlier, $\td$ is smaller than the effective dimension of a tree-structured function and the effective dimension is usually much smaller than $d$.

\subsubsection{Adapting the Kernel Hyper-Parameters}

The GP-UCB algorithm proposed in \cite{srinivas2010} is shown to have an explicit sublinear regret bound $R_t =\mathcal{O}(\sqrt{t\beta_t \gamma_t})$ with high probability for commonly used covariance functions, including the squared exponential kernel and the Mat\'{e}rn kernel.
Here $\gamma_t$ is the maximum mutual information gain after $t$ rounds and is defined to be:
\begin{equation}
\label{eq:maximum-mutual-definition}
    \gamma_t \coloneqq \max_{A \subset \mathcal{X}, |A| = t} I(\bm{y}_A; f),
\end{equation}
where $I(\bm{y}_A; f) = 1 / 2 \log |\bm{I} + \sigma^{-2} \bm{K}_A|$ is the mutual information between observations in $A$ and prior of $f$ under a GP model, $\bm{K}_A$ is the Gram matrix $[ k_{\theta}(\bm{x}, \bm{x}') ]_{\bm{x}, \bm{x}' \in A}$ and is dependent on kernel $k$'s hyperparameters $\theta$.
The squared exponential kernel is defined as $k(\bm{x}, \bm{x}') = \exp (- \frac{1}{2 \theta^2} \norm{\bm{x} - \bm{x}'}_2^2)$, where $\theta$ is the lengthscale parameter. The Mat\'{e}rn kernel is defined as $k(\bm{x}, \bm{x}') = \frac{2^{1 - \nu}}{\Gamma(\nu)} r^{\nu} B_{\nu}(r)$, where $r = \frac{\sqrt{2 \nu}}{\theta} \norm{\bm{x} - \bm{x}'}_2$, $B_{\nu}$ is the modified Bessel function with $\nu > 1$, $\Gamma$ is the gamma function.

One practical issue is \cite{srinivas2010} requires information about the RKHS norm bound of $f$ and the correct hyperparameters of the GP's covariance function, which are usually unavailable.
In this work, we follow the adaptive GP-UCB algorithm (A-GP-UCB) proposed in~\cite{berkenkamp2019}, which can adapt the norm bound of an RKHS and hyperparameters of a covariance function. The adapting mechanism in \cite{berkenkamp2019} is by setting:
\begin{align}
   \beta_t^{1/2} &= B_t + 4 \sigma \sqrt{ I_{\theta_t}(\bm{y}_t; f) + 1 + \ln(1/\delta) } \label{eq:beta-setting} \\
   B_t &= b(t) g(t)^{d} B_0, \quad \theta_t = \theta_0 / g(t) \label{eq:scaling-normbound-lengthscales}
\end{align}
where $\theta_0$ and $B_0$ are initial guess for the lengthscales of kernel $k$ and RKHS norm bound of $f$ respectively, $g(t): \mathbb{N} \to \mathbb{R}_{>0}$ and $b(t): \mathbb{N} \to \mathbb{R}_{>0}$, with $b(0)=g(0)=1$ are monotonically increasing functions. 

\begin{figure}[h]
\centering
\includegraphics[width=0.99\columnwidth]{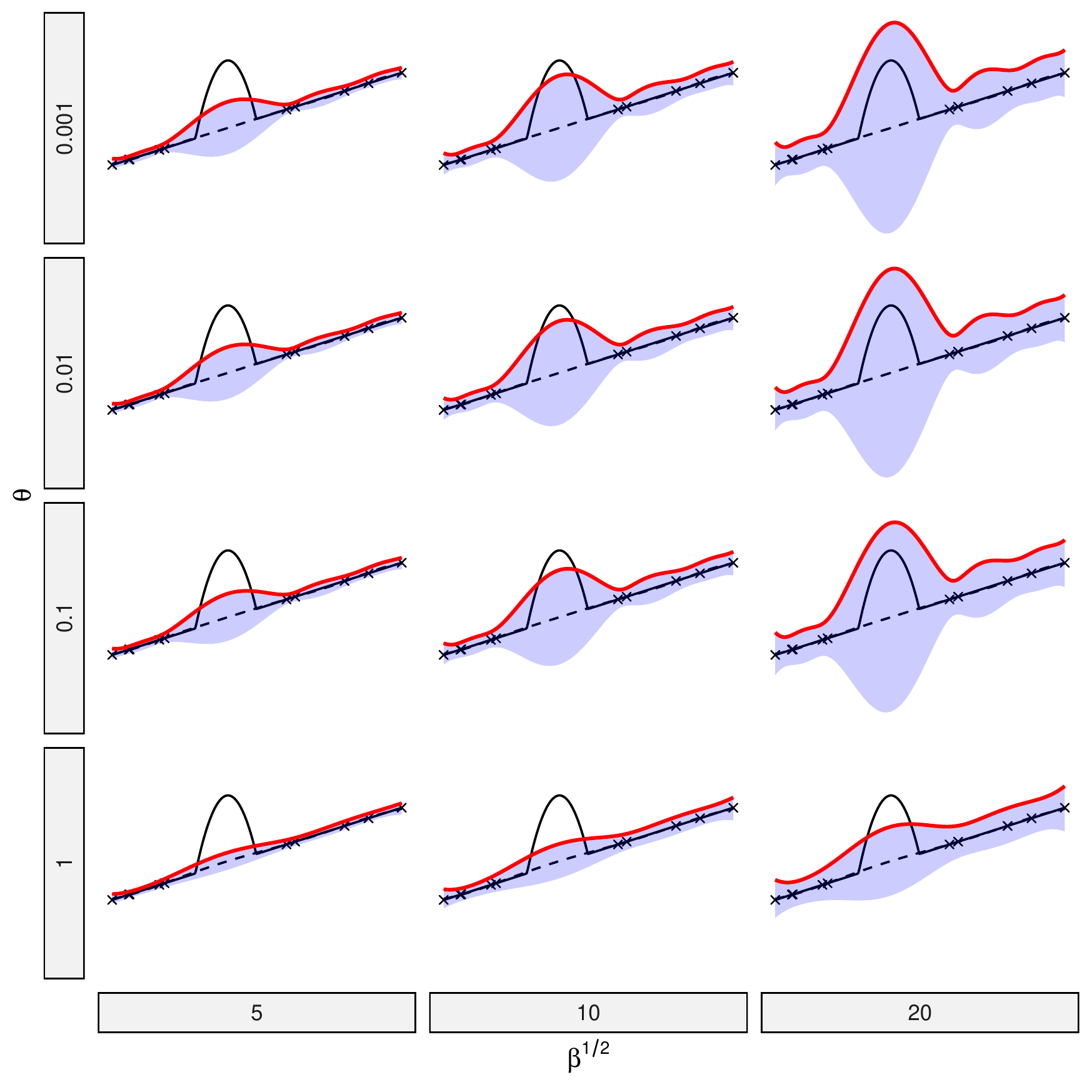}
\caption{Misspecified lengthscales and norm bound cause BO terminate too early. Scaling hyperparameters expands the function class and results in additional exploration. $\beta^{1/2}$ is a proxy to the RKHS bound $B$ and the relationship between $\beta^{1/2}$ and $B$ is shown in \Cref{eq:beta-setting}. 
Crosses indicate the observations so far, $\theta$ shown on the left side indicates the lengthscale of an exponential squared kernel, the dashed line is the posterior mean $\mu(x)$ of a GP model using this covariance function, the shaded area is the confidence interval $\mu(x) \pm \beta^{1/2} \sigma(x)$, where $\sigma(x)$ is the posterior standard deviation,
the solid line indicates the true function, the red line is the upper bound of this confidence interval, which is the GP-UCB acquisition function.}
\label{fig:lengthscale-beta-effect}
\end{figure}

The motivation of adapting the norm bound and the lengthscales is shown in \Cref{fig:lengthscale-beta-effect}. 
As pointed out in \cite{berkenkamp2019}, it is always possible that a local optimum is located in a local bump area and zero observation falls into that region. 
GP-UCB with misspecified lengthscale $\theta$ and RKHS bound $B$ cannot fully explore the input space and this causes BO to terminate too early. 
For example, in the first column of \Cref{fig:lengthscale-beta-effect}, if we maximize the GP-UCB acquisition function (the red line), the algorithm will terminate prematurely and we will never have the chance of exploring the region containing the global maximum because of misspecified hyperparameters. 
Intuitively, decreasing the lengthscales of a covariance function and increasing the norm bound have the effect of expanding the function class, which results in additional exploration, and eventually the true function is contained in the considered space.

\subsubsection{Bounds of Cumulative Regrets}

A high-level summary of this section is as follows. 
Firstly, we use \Cref{th:berkenkamp2019-1} to bound the cumulative regret of our proposed Add-Tree covariance function in terms of the norm bound and the maximum information gain. 
Secondly, we show a tree-structured function has bounded RKHS norm in an RKHS with an Add-Tree kernel under suitable assumptions using \Cref{prop:RKHS-bound-addtree-function}. 
Thirdly, to upper bound this information gain quantity, we use \Cref{th:srinivas2010-8} to connect it with the tail sum of the operator spectrum of an Add-Tree covariance function. 
Fourthly, we derive the upper bound of this tail sum when each covariance function defined on the vertices of $\mathcal{T}$ is a squared exponential kernel or a Mat\'{e}rn kernel using \cref{prop:weyl-extension} and \Cref{th:bounds-kernels}.
Finally, we combine all these together to obtain the cumulative regret of an Add-Tree covariance function.

\begin{theorem}[Theorem 1 in \cite{berkenkamp2019}]
\label{th:berkenkamp2019-1}
Assume that $f$ has bounded \textrm{RKHS} norm $\norm{f}_{k_\theta}^2 \le B$ in a \textrm{RKHS} that is parametrized by a stationary kernel $k_{\theta}(\bm{x}, \bm{x}')$ with unknown lengthscales $\theta$. Based on an initial guess, $\theta_0$ and $B_0$, deﬁne monotonically increasing functions $g(t) > 0$ and $b(t) > 0$ and run \textrm{A-GP-UCB} with $\beta_t^{1/2} = b(t) g(t)^d B_0  + 4 \sigma \sqrt{I_{\theta_t}(\bm{y}_t;\bm{f}) + 1 + \ln (1/\delta) }$ and GP legnthscales $\theta_t = \theta_0 / g(t)$. Then, with probability at least $1-\delta$, we obtain a regret bound of:
\begin{align}
\label{eq:th-berkenkamp2019-1}
    R_t \le& 2 B \max \left(  g^{-1}(\max_i \frac{[\theta_0]_i}{[\theta]_i}), b^{-1}(\frac{B}{B_0})   \right)  \nonumber \\
    &+\sqrt{C_1 t \beta_t I_{\theta_t}(\bm{y}_t;f)}
\end{align}
where $I_{\theta_t}$ is the mutual information based on a GP model with lengthscales $\theta_t$ and $C_1 = 8 / \log (1 + \sigma^{-2})$.
\end{theorem}

Before we turn to our main theoretical results, we recall and extend some existing notation. 
Let $f_{\mathcal{T}}$ be a tree-structured function defined in \Cref{def:TreeStructuredFunction} and $f_{\mathcal{T}}$ satisfies the additive decomposition in \Cref{eq:additive-assumption}. 
Let $M$ be the height of $\mathcal{T}$, $v_{ij}$ be the $j$-th vertex on the $i$-th level of $\mathcal{T}$, where $i \in \{ 1,\dots,|P| \}, j \in \{ 1,\dots, M\}$. Let $k_{ij}$ be a convariance function defined on the continuous variables appearing at vertex $v_{ij}$, $\mathcal{H}_{ij}$ be an RKHS with kernel $k_{ij}$, $f_{ij}$ be a function defined on the continuous variables on vertex $v_{ij}$ and assume $f_{ij} \in \mathcal{H}_{ij}$. Let $\td$ be the highest dimension of functions in $\{ f_{ij} \}$, $f_i \coloneqq \sum_{j=1}^{h_i} f_{ij}$ denote the sum of functions defined on the $i$-th path of $\mathcal{T}$, and let $k_i$ be the sum of all convariance functions on the $i$-th path of $\mathcal{T}$. We note the collection $\{ f_{ij} \}$ is a multiset, since different paths can share variables. 

To utilize \Cref{th:berkenkamp2019-1}, we need to ensure a tree-structured function $f_{\mathcal{T}}$ has bounded RKHS norm in an RKHS with an Add-Tree covariance function $k_{\mathcal{T}}$.

\begin{proposition}
\label{prop:RKHS-bound-addtree-function}
Assume every $f_{ij}$ has bounded RKHS norm $\norm{f_{ij}}_{k_{ij}}^{2} \le \tilde{B}$ in an RKHS with a kernel $k_{ij}$, and let $k_{\mathcal{T}}$ be an Add-Tree covariance function constructed from $k_{ij}$, then $f_\mathcal{T}$ has bounded RKHS norm in an RKHS with a kernel $k_{\mathcal{T}}$:
\begin{equation}
    \norm{f_\mathcal{T}}_{k_{\mathcal{T}}}^2 \le M \tilde{B}.
\end{equation}
\end{proposition}

\begin{theorem}[Theorem 8 in~\cite{srinivas2010}]
\label{th:srinivas2010-8}
Suppose that $D \subset \mathbb{R}^d$ is compact, and $k(\bm{x},\bm{x}')$ is a covariance function for which the additional assumption of Theorem 2 in \cite{srinivas2010} holds. Moreover, let $B_{k}(T_{*}) =\sum_{s > T_{*}} \lambda_{s}$, where $\{ \lambda_s \}$ is the operator spectrum of $k$ with respect to the uniform distribution over $D$. 
Pick $\tau > 0$, and let $n_{T}=C_{4}T^{\tau} \log T$, with $C_4$ depends only on $\tau$ and $D$. Then the following bound holds true:
\begin{align}
\gamma_{T} &\leq \frac{1/2}{1-e^{-1}} \max_{r \in \{ 1,\dots,T \} } \bigg( T_{*} \log (\frac{r n_T}{\sigma^2})    \nonumber \\
    &+ C_4 \sigma^2 (1- \frac{r}{T})( B_{k}(T_*) T^{\tau + 1} + 1 ) \log T \bigg)    \nonumber \\
    &+ \mathcal{O}(T^{1-\tau/d})
    \label{eq:srinivas2010-max-gain}
\end{align}
for any $T_* \in \{ 1,\dots, n_T \}$.
\end{theorem}

Using \Cref{th:srinivas2010-8}, we can bound the maximum information gain in terms of $B_k(T_*)$, the tail sum of the operator spectrum of a kernel $k$. Our first main theoretical result is shown in the following \Cref{th:bounds-kernels}. Comparing with the results in \cite{berkenkamp2019}, our bounds of $\gamma_t$ for two commonly used covariance functions have a linear dependence with the total dimension $d$ and exponential dependence with $\td$, which is usually much smaller than $d$ in a tree-structured function.

\begin{proposition}
\label{th:bounds-kernels}
Let $k_{\mathcal{T}}$ be an Add-Tree covariance function defined in 
\Cref{def:Add-Tree} and is parametrized by lengthscales $\theta$. Scaling lengthscales of $k_{\mathcal{T}}$ and norm bound at iteration $t$ by \Cref{eq:scaling-normbound-lengthscales}.

\begin{itemize}
    \item If every covariance function $k_{ij}$ defined on vertex $v_{ij}$ is a squared exponential kernel, then 
    \begin{equation*}
        \gamma_t = \mathcal{O} (d \tilde{d}^{\tilde{d}} (\log t)^{\tilde{d}+1} g(t)^{2\tilde{d}}) 
    \end{equation*}
    
    \item If every covariance function $k_{ij}$ defined on vertex $v_{ij}$ is a Mat\'{e}rn kernel, then 
    \begin{equation*}
       \gamma_t = \mathcal{O}(M^2 g(t)^{2 \nu+\td} (\log t) t^{\frac{\td (\td +1)}{2 \nu + \td (\td +1)}}) 
    \end{equation*}
\end{itemize}
\end{proposition}

\begin{proposition}
\label{coro:cumulative-regret}
Under the assumptions of \Cref{th:bounds-kernels}, $b(t)$ and $g(t)$ grow unbounded, then we obtain the following high-probability bounds using our proposed Add-Tree covariance function:

\begin{itemize}
    \item Squared exponential kernel: 
    \begin{align*}
        R_t =& \mathcal{O}\bigg( \sqrt{t} g(t)^{2 \td} b(t) M \sqrt{d \td^{\td} (\log T)^{\tilde{d}+1}} \\
        &\quad+ \sqrt{t} g(t)^{2 \td} d \td^{\td} (\log T)^{\td+1} \bigg)
    \end{align*}
    
    \item Mat\'{e}rn kernel:
    \begin{align*}
        R_t =& \mathcal{O} \bigg( M^2 b(t) \sqrt{t g(t)^{2 \nu + 3 \td} (\log t) t^{1 - \tau \td}} \\
        &\quad+ M^2 g(t)^{2 \nu + \td} (\log t) t^{1 - \tau \td} \sqrt{t}  \bigg)
    \end{align*}
    where $\tau=2 \nu \td /(2 \nu + \td (1+\td))$.
\end{itemize}

\end{proposition}

\Cref{coro:cumulative-regret} should be compared with the results given in \cite{berkenkamp2019}, where bounds of $R_t$ have an exponential dependence with $d$, while our bounds have an exponential dependence with $\td$. This improvement is a result of the assumed additive structure.
Since we are adapting the norm bound and the lengthscale, and we use the additive assumption, our regret bounds in \Cref{coro:cumulative-regret} are not directly comparable to the results stated in \cite{srinivas2010}. 
Fortunately, insights can be gained by comparing the forms of \Cref{coro:cumulative-regret} and the results in \cite{srinivas2010}. In the initial stage, the true function is not contained in the current space, since the misspecified norm bound $B_t$ or the RKHS space associated with $k_{\theta_t}$ are too small. In this case, we can bound the regret by $2B$\cite{berkenkamp2019}. Since $g(t)$ and $b(t)$ grow unbounded, $B_t$ can be larger and $\theta_t$ can be smaller than their optimal values. Consider the case for a squared exponential kernel, the additional factor $g(t)^{2 \td}b(t)$ in the first term leads to additional exploration and is the price we pay for our ignorance of the true norm bound and the lengthscale parameter.


\subsection{Practical Considerations}
\label{sec:practical-consideration}

Although \Cref{coro:cumulative-regret} holds for $b(t)$ and $g(t)$ growing unbounded, what we really want is an asymptotically no-regret algorithm:
\begin{equation*}
    \lim_{t \to \infty} R_t / t = 0
\end{equation*}
To achieve this goal, we need to choose $b(t)$ and $g(t)$ from a narrower class and make $\sqrt{\beta_t \gamma_t}$ grow slower than $\sqrt{t}$. Several methods to choose $b(t)$ and $g(t)$ are given in \cite{berkenkamp2019}. In brief, (i) we first select a reference regret, reflecting our willingness to perform additional exploration. (ii) Then we match an estimator of the cumulative regret with this reference regret. (iii) finally we compute the $g(t)$ and $b(t)$.  

Two natural questions are whether we can test the validity of the additive assumption, and what to do if the additive assumption doesn't hold. First, the cornerstone of BO is a probabilistic regression model and BO inherits all the limitations of this regression model. That means there is little we can do in a high-dimensional space, even with hundreds of thousands of observations \cite{gyorfi2002}, unless we use dimension reduction techniques.  
It is thus generally unreliable to test if the additive assumption is satisfied in BO settings, because we are almost always in a low-data regime. Additionally, automatically discovering an additive structure based on a few hundreds of observations in a high-dimensional space could lead to serious overfitting. 
The second question has nothing to do with the inference procedure proposed in this work, but rather a model assumption problem. It is always possible to fit a GP model using our proposed Add-Tree covariance function based on the observations collected in the initial phase of BO. If the mean squared error is sufficient, we can continue the BO.


\section{EXPERIMENTS}
\label{sec:experiment}

In this section, we present empirical results using four sets of experiments. 
To demonstrate the efficiency of our proposed Add-Tree covariance function, we first optimize a synthetic function presented in~\cite{jenatton2017} in \Cref{sec:synthetic-function}, comparing with SMAC~\cite{hutter2011}, TPE~\cite{bergstra2011}, random search~\cite{bergstra2012}, standard GP-based BO from GPyOpt~\cite{gpyopt2016}, and the semi-parametric approach proposed in~\cite{jenatton2017}. To facilitate our following description, we refer to the above competing algorithms as \textbf{smac}, \textbf{tpe}, \textbf{random}, \textbf{gpyopt}, and \textbf{tree} respectively. Our approach is referred as \textbf{add-tree}. 
To verify our Add-Tree covariance function indeed enables sharing between different paths, we compare it with independent GPs in a regression setting showing greater sample efficiency for our method. 
We then apply our method to compress a three-layer fully connected neural network, VGG16 and ResNet50 in \Cref{sec:model-compression} and \Cref{sec:model-pruning}.
Finally, in \Cref{sec:resnet20-nas}, we search for activation functions of ResNet20 and their corresponding hyper-parameters.
Experimental results show our method outperforms all competing algorithms. 

For all GP-based BO, including \textbf{gpyopt}, \textbf{tree} and \textbf{add-tree}, we use the squared exponential covariance function. To optimize the parameters of Add-Tree, we maximize the marginal log-likelihood function of the corresponding GP and set $\theta_t = \min(\theta_t^{\text{MAP}}, \theta_0 / g(t))$ following \cite{berkenkamp2019}. 
As for the numerical routine used in fitting the GPs and optimizing the acquisition functions, we use multi-started L-BFGS-B, as suggested by~\cite{kim2019}.

In our experiments, we use official implementations of competing algorithms. For \textbf{tpe}, we use Hyperopt~\cite{bergstra2013}. For \textbf{smac}, we use SMAC3\footnote{\url{https://github.com/automl/SMAC3}}.
We note the original code for~\cite{jenatton2017} is unavailable, thus we have re-implemented their framework to obtain the results presented here. There are several hyper-parameters in their algorithm which are not specified in the publication.  To compare fairly with their method, we tune these hyper-parameters such that our implementation has a similar performance on the synthetic functions to that reported by \cite{jenatton2017}, and subsequently fix the hyper-parameter settings in the model compression task in \Cref{sec:model-compression}.

\subsection{Synthetic Experiments}
\label{sec:synthetic-function}

\begin{figure}[h]
\centering
\resizebox{0.95\columnwidth}{!}{
\begin{tikzpicture}[
    sibling distance        = 12em,
    level distance          = 5em,
    edge from parent/.style = {draw, -latex},
    every node/.style       = {font=\footnotesize},
    sloped,
    ]
  \node [root] {$x_1$}
    child {
      node [root] {$x_2,r_8$}
        child {
          node [env] {$x_4^2 + 0.1 + r_8$}
          edge from parent
            node[above] {0}
        }
        child {
          node [env] {$x_5^2 + 0.2 + r_8$}
          edge from parent
            node [above] {1}
        }
      edge from parent
        node [above] {0}
    }
    child {
      node [root] {$x_3,r_9$}
        child {
          node [env] {$x_6^2 + 0.3 + r_9$}
          edge from parent
            node[above] {0}
        }
        child {
          node [env] {$x_7^2 + 0.4 + r_9$}
          edge from parent
            node [above] {1}
        }
      edge from parent
        node [above] {1}
    };
\end{tikzpicture}
}
\caption{A synthetic function from~\cite{jenatton2017}. The dimension of this function is $d=9$ and the effective dimension at any leaf is $2$. }
\label{fig:synthetic-function-jenatton2017}
\end{figure}
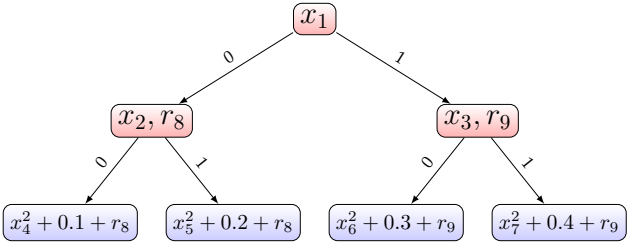

In our first experiment, we optimize the synthetic tree-structured function depicted in \Cref{fig:synthetic-function-jenatton2017}. This function is originally presented in~\cite{jenatton2017}. In this example, continuous variables include $x_4,x_5,x_6,x_7,r_8,r_9$, and $r_8,r_9$ of them are shared variables. $x_4,x_5,x_6,x_7$ are defined in $[-1,1]$ and $r_8,r_9$ are bounded in $[0,1]$. Categorical variables include $x_1,x_2,x_3$ and they are all binary.

\begin{figure*}[h]
\centering
\subfloat[Synthetic function optimization]{
\includegraphics[width=0.32\textwidth]{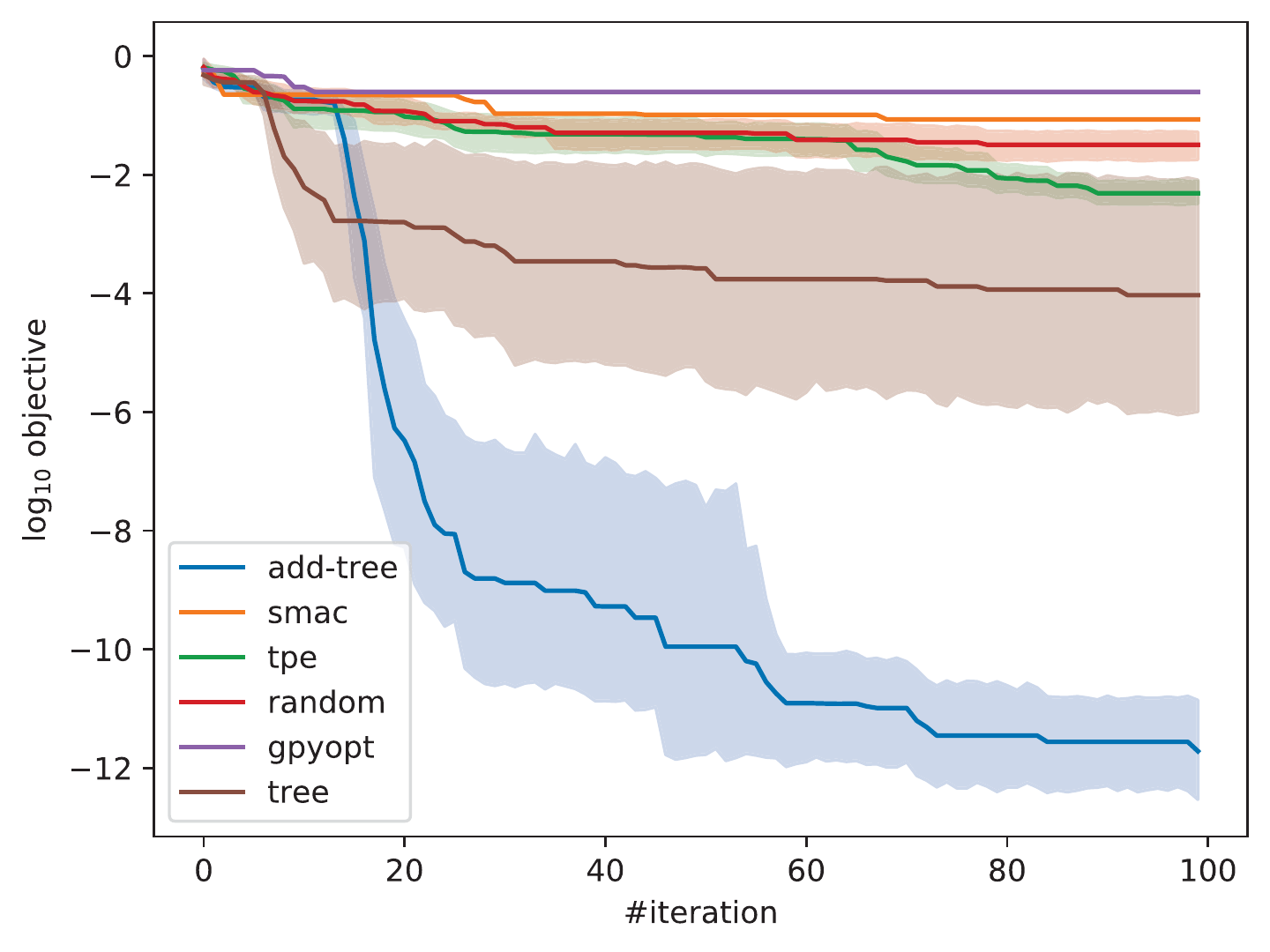}
\label{fig:exp:synthetic-optimization-comparison}}
\hfil
\subfloat[Synthetic function regression]{
\includegraphics[width=0.32\textwidth]{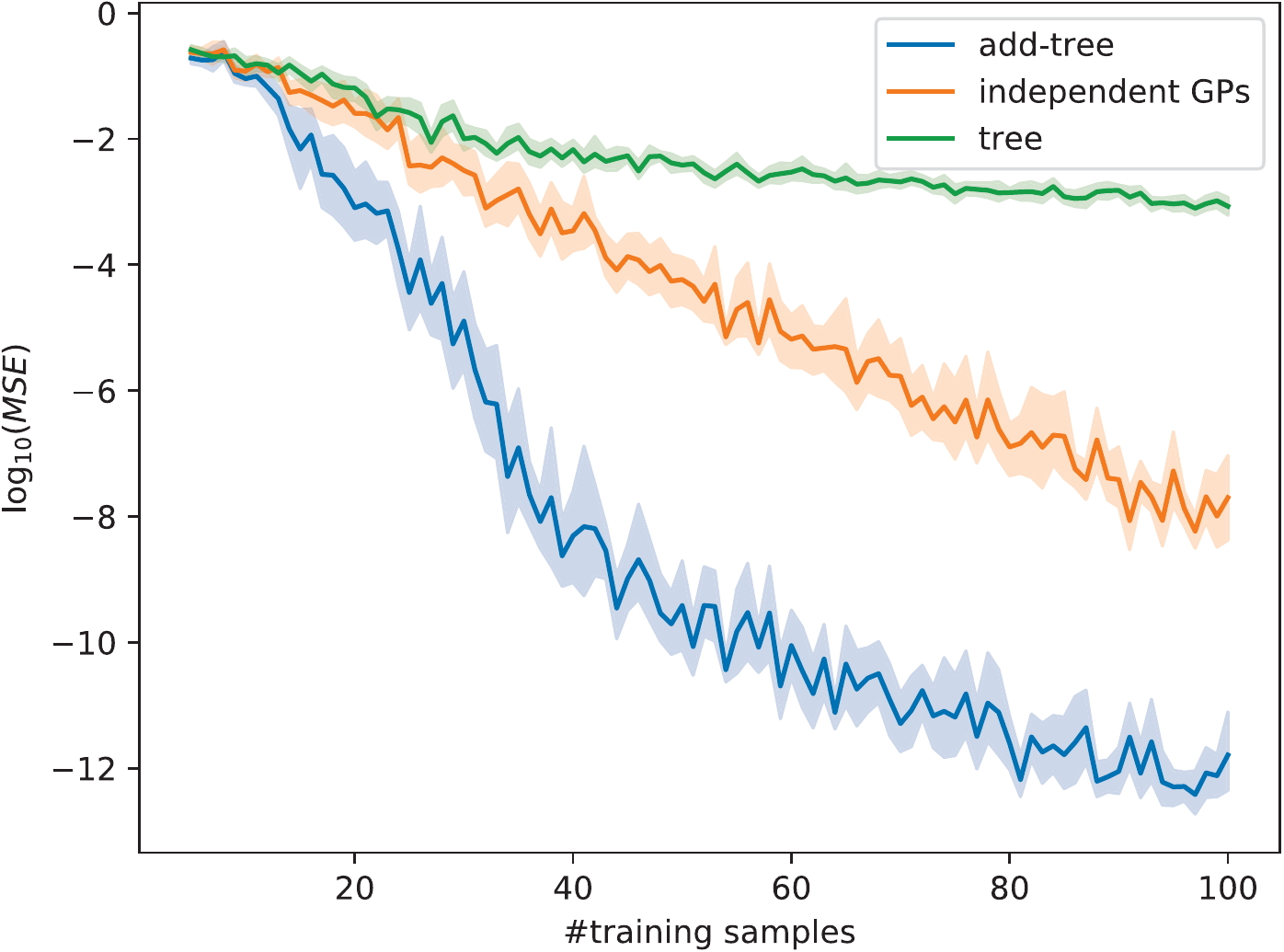}
\label{fig:regression-comparison}}
\hfil
\subfloat[FC3 compression optimization]{
\includegraphics[width=0.32\textwidth]{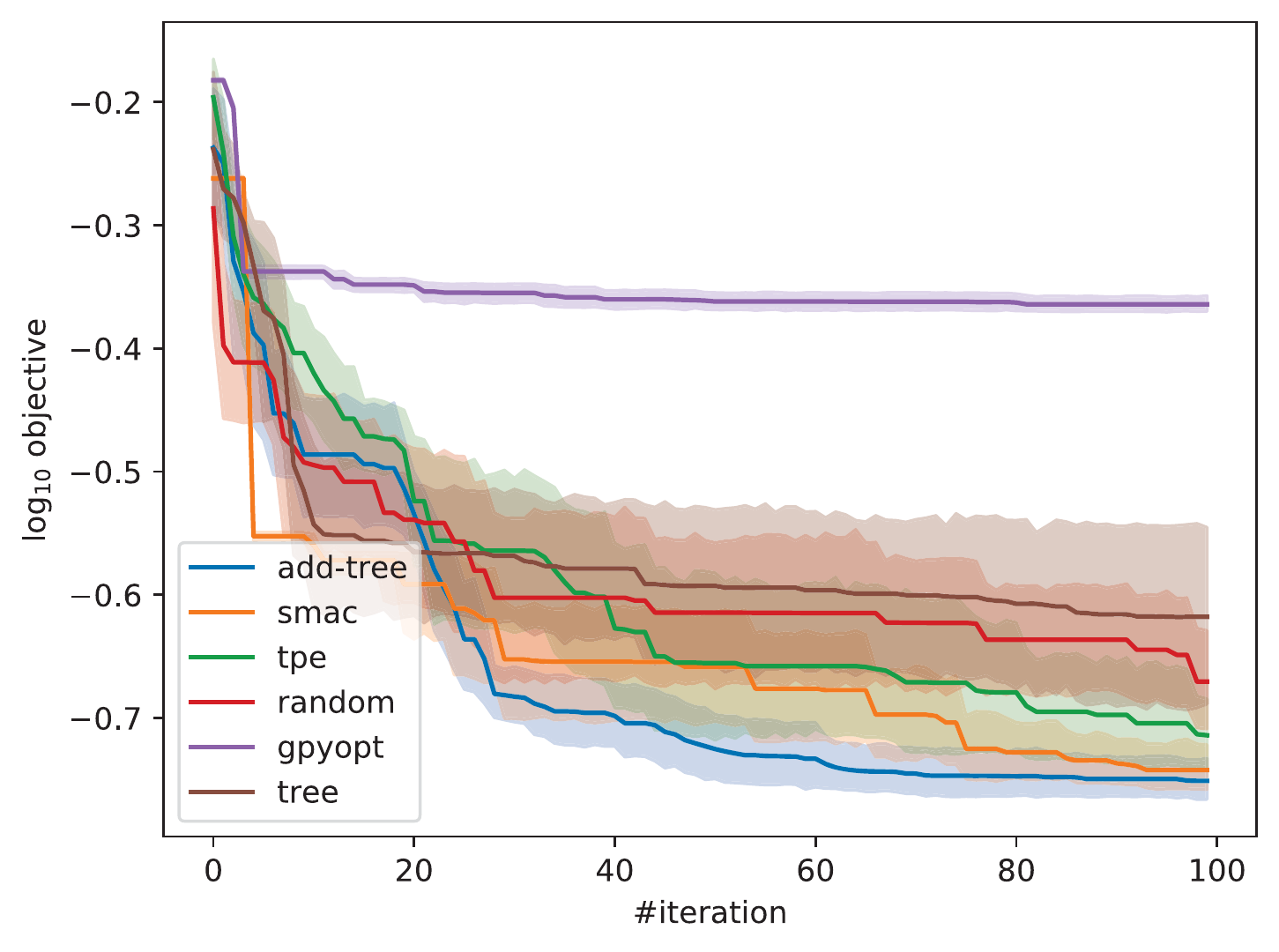}
\label{fig:exp:fc3-compression-comparison} }
\caption{Comparison results on minimizing a synthetic function and compressing FC3 using different algorithms. The solid line is the incumbent mean of 10 independent runs, and the shaded region is twice the standard deviation (95 \% confidence interval for mean) of these 10 runs. \Cref{fig:exp:synthetic-optimization-comparison} and \Cref{fig:exp:fc3-compression-comparison} shows six different algorithms on optimizing the synthetic function and compressing a simple 3-layer neural network respectively. \Cref{fig:regression-comparison} shows the performance comparison in a regression setting, indicating our proposed Add-Tree covariance function indeed enables information sharing between different paths. }
\label{fig:syntetic-regression-compression-exp-results}
\end{figure*}

\Cref{fig:exp:synthetic-optimization-comparison} shows the optimization results of six different methods. The x-axis shows the iteration number and the y-axis shows the $log_{10}$ distance between the best minimum achieved so far and the known minimum value, which in this case is $0.1$. It is clear from \Cref{fig:exp:synthetic-optimization-comparison} that our method has a substantial improvement in performance compared with other algorithms. After 60 iterations, the $\log_{10}$ distance of~\textbf{tree} is still higher than -4, while our method can quickly obtain a much better performance in less than 20 iterations. Interestingly, our method performs substantially better than independent GPs, which will be shown later, while in~\cite{jenatton2017}, their algorithm is inferior to independent GPs. We note \textbf{gpyopt}\footnote{GPyOpt  \cite{gpyopt2016} is a state-of-the-art open source Bayesian optimization software package with support for categorical variables.} performs worst, and this is expected (recall \Cref{subsec:bo}). \textbf{gpyopt} encodes categorical variables using a one-hot representation, thus it actually works in a space whose dimension is $d'=d+c=12$, which is relatively high considering we have less than 100 data points. In this case, \textbf{gpyopt} behaves like random search, but in a 12-dimensional space instead of the 9-dimensional space of a na\"{i}ve random exploration.


To show that Add-Tree allows efficient information sharing across different paths, we compare it with independent GPs and \textbf{tree} in a regression setting and results are shown in \Cref{fig:regression-comparison}. 
The training data is generated from the synthetic function in \Cref{fig:synthetic-function-jenatton2017}: categorical values are generated from a Bernoulli distribution with $p=0.5$, continuous values are uniformly generated from their domains. The x-axis shows the number of generated training samples and the y-axis shows the $\log_{10}$ of Mean Squared Error (MSE) on a separate randomly generated data set with 50 test samples. 
From \Cref{fig:regression-comparison}, we see that Add-Tree models the response surface better even though independent GPs have more parameters. For example, to obtain a test performance of $10^{-4}$, Add-Tree needs only 24 observations, while independent GPs require 44 data points. If we just look at the case when we have 20 training samples, the absolute MSE of independent GPs is $10^{-1}$, while for Add-Tree, it is $10^{-3}$. The reason for such a huge difference is when training data are rare, some paths will have few data points, and Add-Tree can use the shared information from other paths to improve the regression model. This property of Add-Tree is valuable in BO settings.


\subsection{Model Compression}
\label{sec:model-compression}

Neural network compression is essential when it comes to deploying models on devices with limited memory and low computational resources. For parametric compression methods, like Singular Value Decomposition (SVD) and unstructured weight pruning (P-L1), it is necessary to tune their parameters in order to obtain the desired trade-off between model size and performance.  Existing publications on model compression usually determine parameters for a single compression method, and do not have an automated selection mechanism over different methods. By encoding this problem using a tree-structured function, different compression methods can now be applied to different layers and this formulation is more flexible than the current literature. 

In this experiment, we apply our method to compress a three-layer fully connected network \textit{FC3} defined in \cite{Ma2019a}. FC3 has 784 input nodes, 10 output nodes, 1000 nodes for each hidden layer and is pre-trained on the MNIST dataset. For each layer, we optimize a discrete choice for the compression method, which is among SVD and P-L1, then optimize either the rank of the SVD or the pruning threshold of P-L1. We only compress the first two layers, because the last layer occupies 0.56\% of total weights. The rank parameters are constrained to be in $[10, 500]$ and the pruning threshold parameters are bounded in $[0,1]$. The tree structure of this problem has a depth of 3, and there is no continuous parameter associated with the root. Following \cite{Ma2019a}, the objective function used in compressing FC3 is: 
\begin{equation}
\label{eq:obj-compression}
    \gamma \mathcal{L}(\tilde{f_\theta}, f^*) + R(\tilde{f_\theta}, f^*) ,
\end{equation}
where $f^*$ is the original FC3, $\tilde{f_\theta}$ is the compressed model using parameter $\theta$, $R(\tilde{f_\theta}, f^*)$ is the compression ratio and is defined to be the number of weights in the compressed network divided by the number of weights in the original network.  $\mathcal{L}(\tilde{f}_\theta, f^*) \coloneqq \mathbb{E}_{x \sim P}(\|\tilde{f}_\theta(x) - f^*(x)\|_2^2)$, where $P$ is an empirical estimate of the data distribution. Intuitively, the $R$ term in \Cref{eq:obj-compression} prefers a smaller compressed network, the $\mathcal{L}$ term prefers a more accurate compressed network and $\gamma$ is used to trade off these two terms. In this experiment, $\gamma$ is fixed to be 0.01, and the number of samples used to estimate $\mathcal{L}$ is 50 following \cite{Ma2019a}. 
\Cref{fig:exp:fc3-compression-comparison} shows the results of our method (Add-Tree) compared with other methods. 
For this experiment, although \textbf{smac}, \textbf{tpe} and \textbf{tree} all choose SVD for both layers at the end, our method converges significantly faster, once again demonstrating our method is more sample-efficient than other competing methods. 
We note \textbf{gpyopt} has the worst performance among all other competing methods in this experiment.


Table~\ref{tab:wilcoxon-test} shows the results of pairwise Wilcoxon signed-rank tests\footnote{We used the Wilcoxon signed rank implementation from scipy.stats.wilcoxon with option (alternative == 'greater').} for the above two objective functions at different iterations.
In Table~\ref{tab:wilcoxon-test}, \textbf{addtree} almost always performs significantly better than other competing methods (significance level $\alpha=0.05$), while no method is significantly better than ours.

\begin{table}[t]
\caption{\label{tab:wilcoxon-test}Wilcoxon Signed-Rank Test}
\centering
\tabcolsep=0.15cm
\begin{tabular}{ccccccc}
\toprule
\Gape[4pt]{\textbf{Experiment}} & \textbf{Iter} &
\textbf{smac} & \textbf{tpe} & \textbf{random} & \textbf{gpyopt} & \textbf{tree} \\
\midrule
 & 40 & 0.003 & 0.030 & 0.005 & 0.003 & 0.018\\
\cmidrule{2-7}
 & 60 & 0.003 & 0.003 & 0.003 & 0.003 & 0.003\\
\cmidrule{2-7}
\multirow{-3}{*}{\centering \makecell{synthetic \\ function}} & 80 & 0.003 & 0.003 & 0.003 & 0.003 & 0.003\\
\cmidrule{1-7}
 & 40 & 0.101 & 0.023 & 0.101 & 0.003 & 0.014\\
\cmidrule{2-7}
 & 60 & 0.037 & 0.018 & 0.011 & 0.003 & 0.008\\
\cmidrule{2-7}
\multirow{-3}{*}{\centering \makecell{model \\ compression}} & 80 & 0.166 & 0.005 & 0.003 & 0.003 & 0.006\\
\bottomrule
\end{tabular}
\end{table}

\subsection{Model Pruning}
\label{sec:model-pruning}

In this section, we perform model pruning on pre-trained VGG16 \cite{simonyan2015} and ResNet50 \cite{he2015} on the CIFAR-10 dataset \cite{Krizhevsky09learningmultiple} and prune the weights of convolutional layers in VGG16 and ResNet50. 
Similar to \Cref{sec:model-compression}, for each layer, the pruning method is automatically chosen among the unstructured weight pruning (P-L1) as in \Cref{sec:model-compression}, and the L2 structured weight pruning (P-L2), which prunes output channels with the lowest L2 norm.
For ResNet50, we don't prune its first block because it occupies only 0.91\% of total convolutional weights. A block in ResNet50 is composed of several bottlenecks, and to prune a bottleneck in a block, we use 4 continuous parameters, corresponding to 4 convolutional layers in this bottleneck. To prune a block, we repeat this process for each bottleneck in this block. In this experiment, these 4 continuous parameters are shared for all bottlenecks in one block, and there are 7 categorical parameters, 56 continuous parameters.\footnote{We use the parameter counting method in SMAC. Using different counting method could lead to different number of categorical parameters, but it doesn't make a difference in practice.}
For VGG16, we group all convolutional layers into 4 blocks in a similar fashion with ResNet50. The first block contains the first 4 convolutional layers and later blocks contain 3 consecutive convolutional layers. We don't prune the first block because it takes up only 1.77\% of total convolutional weights. To prune each block, we use 3 continuous parameters, and there are 7 categorical parameters, 42 continuous parameters in this problem. The tree structure of pruning VGG16 has a depth of 4, and each node in this tree except the root is associated with a 3-dimensional continuous parameter. The tree of pruning ResNet50 has a similar structure to the tree of pruning VGG16, except each node except the root is associated with a 4-dimensional continuous variable. 

In \Cref{sec:synthetic-function} and \Cref{sec:model-compression}, it has been shown \textbf{gpyopt} doesn't perform well in high-dimensional problems, and \textbf{tree} isn't competitive with other algorithms, even worse than \textbf{random} in a model compression problem. Thus in this section, we omit the comparison with these two algorithms. 

The objective function used in this experiment is a trade-off between the model sparsity and the top-1 accuracy on a separate validation set: $S(\tilde{f_\theta}, f^*) + T(\tilde{f_\theta})$, where $S(\tilde{f_\theta}, f^*) = 1 - R(\tilde{f_\theta}, f^*)$, $T(\tilde{f_\theta})$ is the top-1 accuracy of the compressed model after finetuning for two epochs. This objective function is bounded above by 2. In a concrete application, it is possible to add a trade-off parameter between $S(\tilde{f_\theta}, f^*)$ and $T(\tilde{f_\theta})$, depending on whether we care about the size or the accuracy of the pruned model. 
We emphasize that the focus of this work is the comparison with other algorithms in conditional parameter spaces, instead of the comparison with other existing pruning methods. 

\begin{figure*}[h]
\centering
\subfloat[VGG16: all iterations]{
\includegraphics[width=0.31\textwidth]{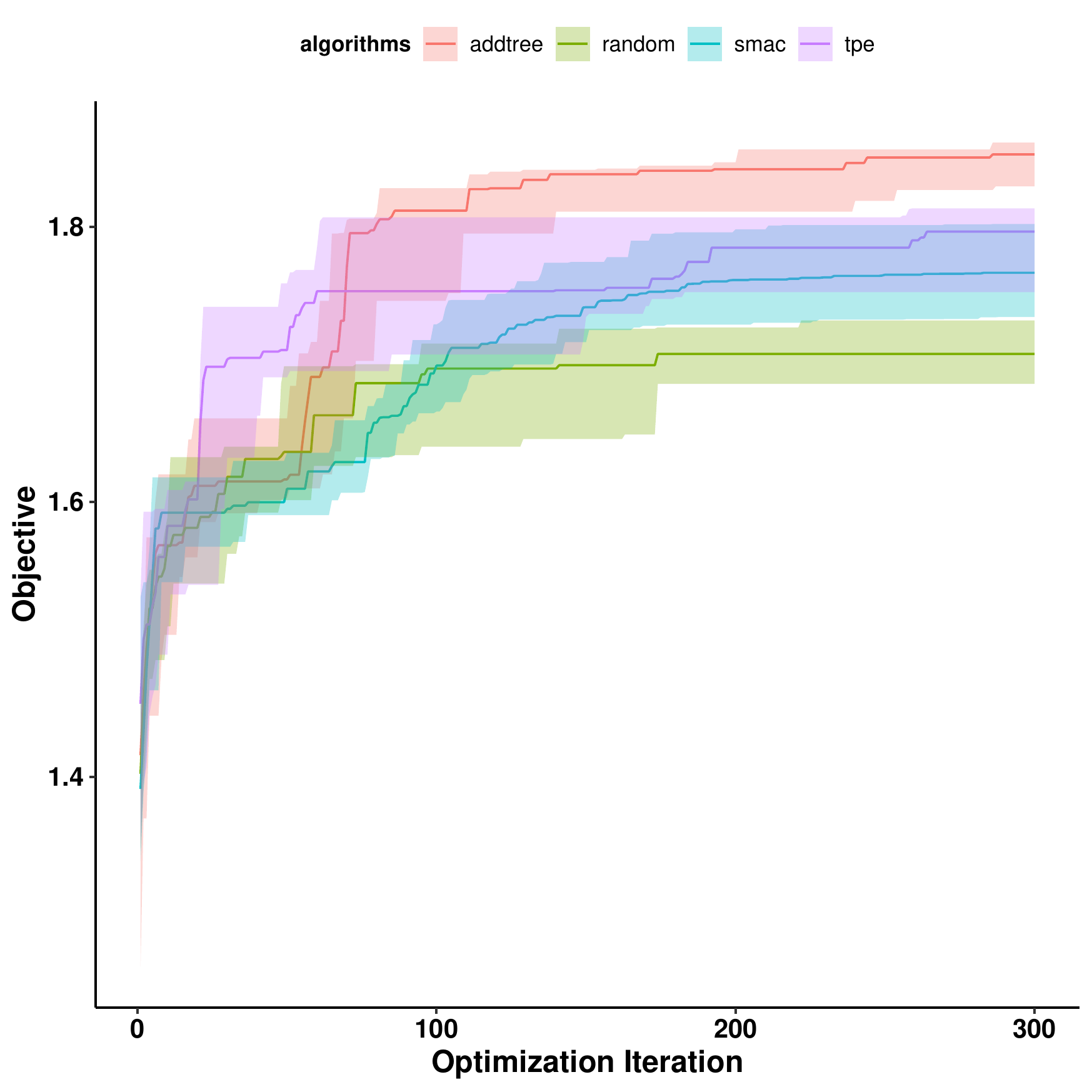}
\label{fig:model-pruning-vgg16-cummax-300-iterations}}
\hfil
\subfloat[VGG16: best of all iterations]{
\includegraphics[width=0.31\textwidth]{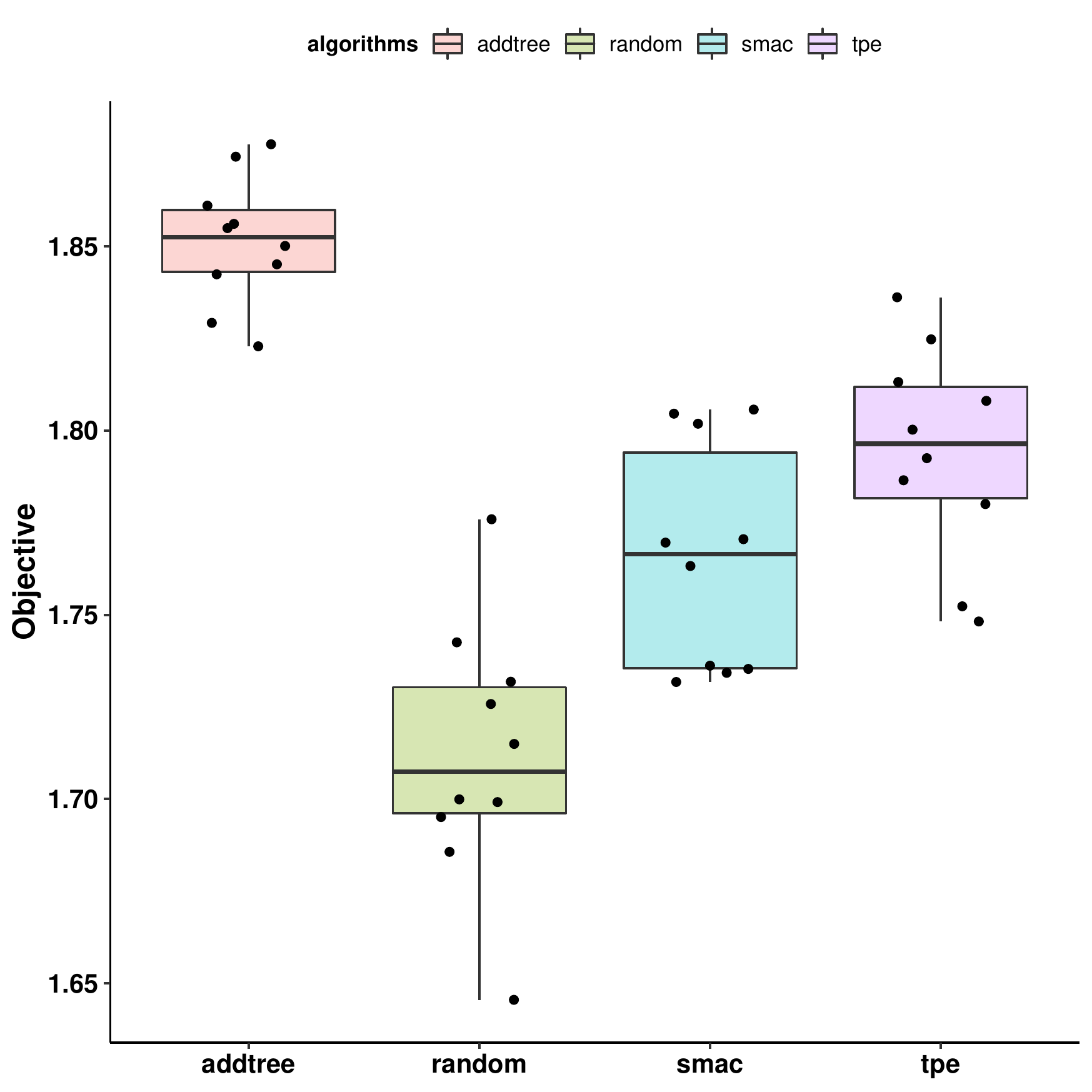}
\label{fig:vgg16-compare-different-algorithms}}
\hfil
\subfloat[VGG16: blockwise compression analysis]{
\includegraphics[width=0.31\textwidth]{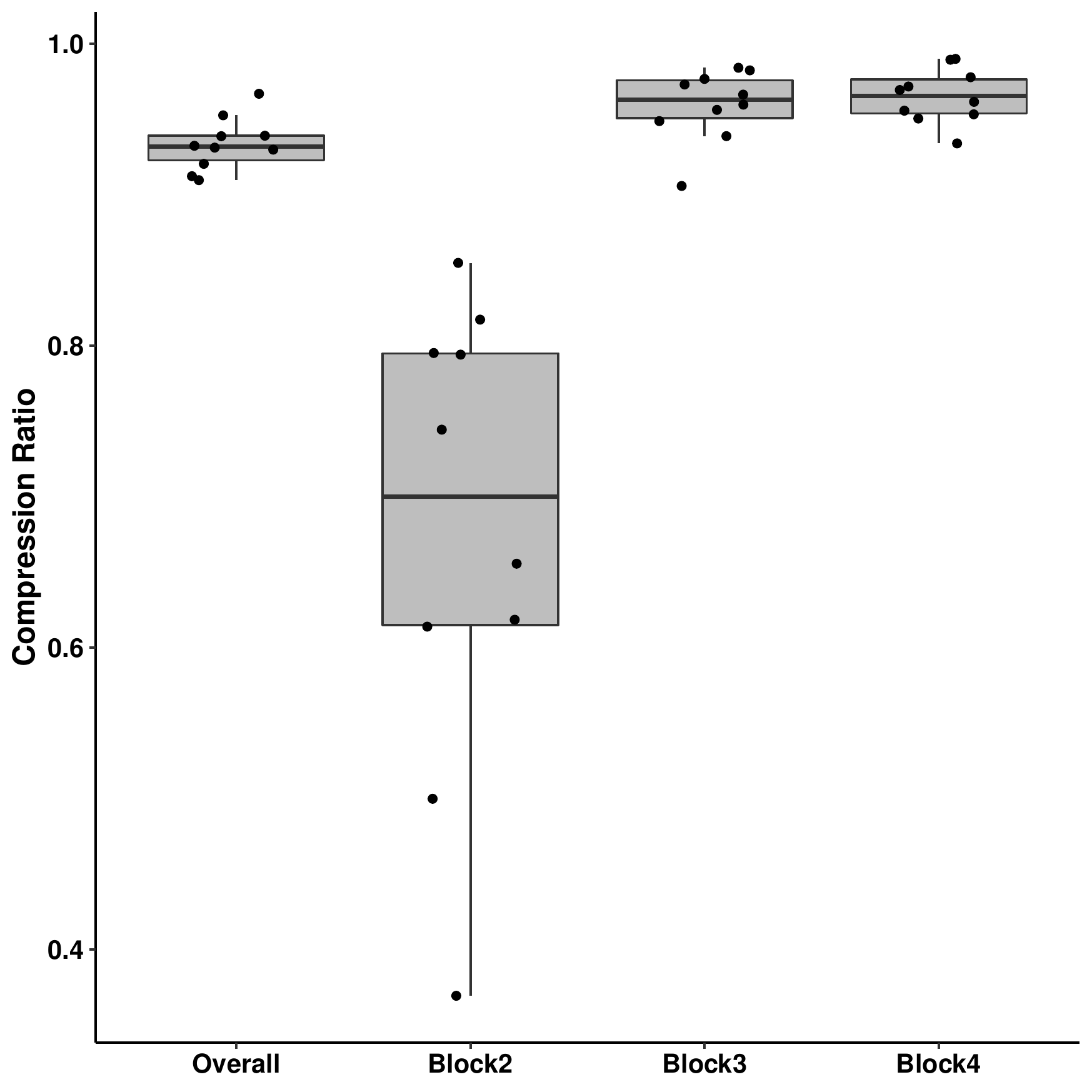}
\label{fig:vgg16-addtree-blockwise-analysis}}

\subfloat[ResNet50: all iterations]{
\includegraphics[width=0.31\textwidth]{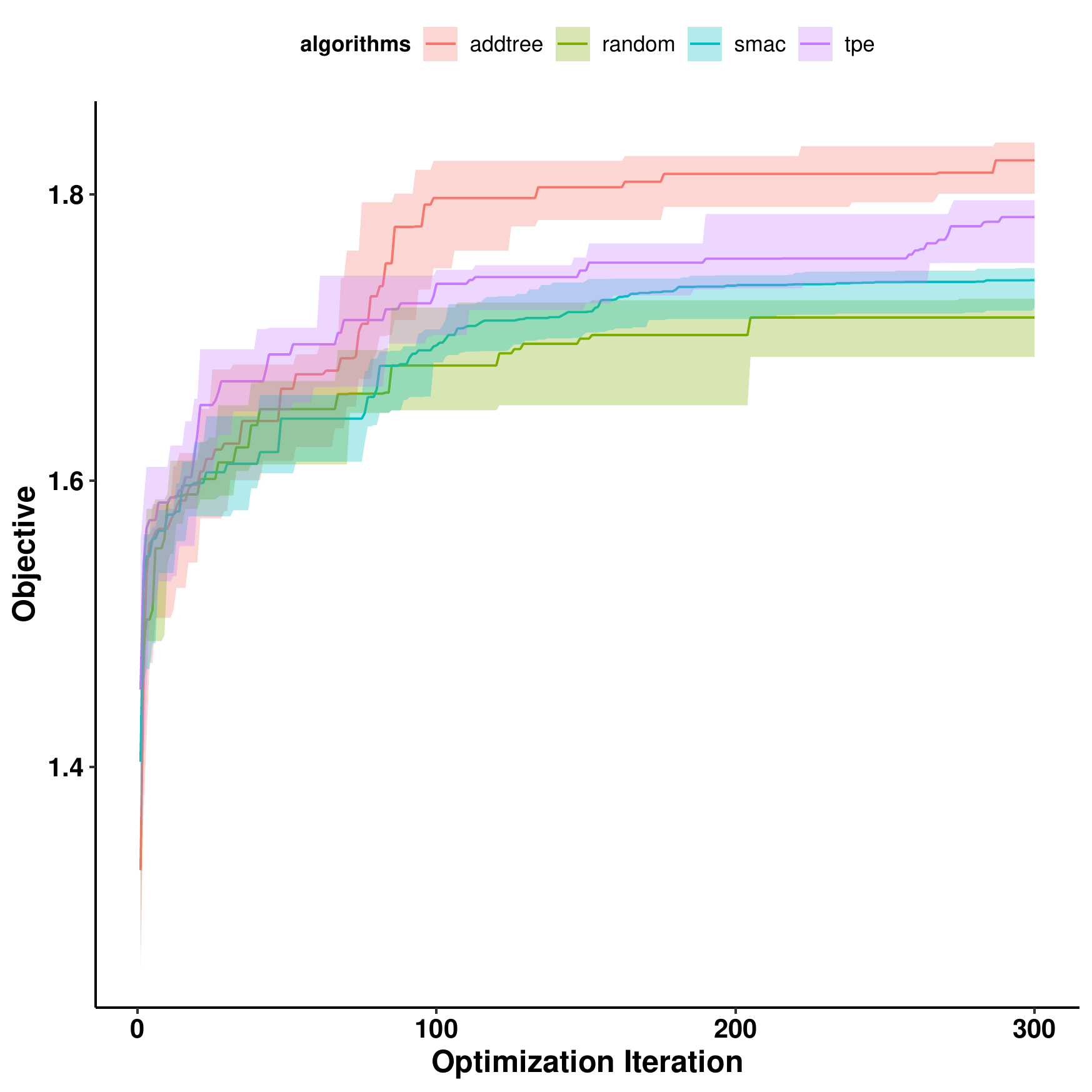}
\label{fig:model-pruning-resnet50-cummax-300-iterations}}
\hfil
\subfloat[ResNet50: best of all iterations]{
\includegraphics[width=0.31\textwidth]{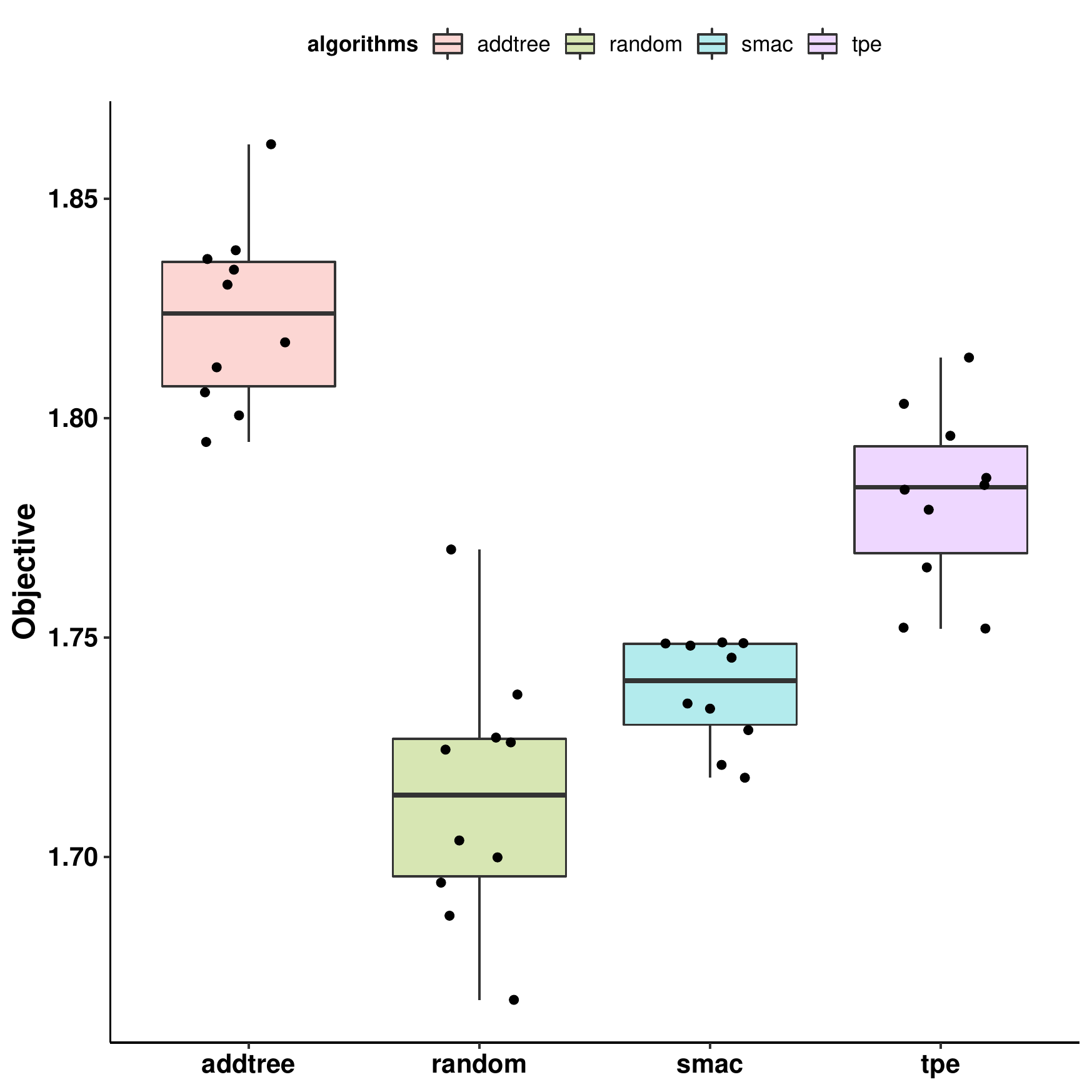}
\label{fig:resnet50-compare-different-algorithms}}
\hfil
\subfloat[ResNet50: blockwise compression analysis]{
\includegraphics[width=0.31\textwidth]{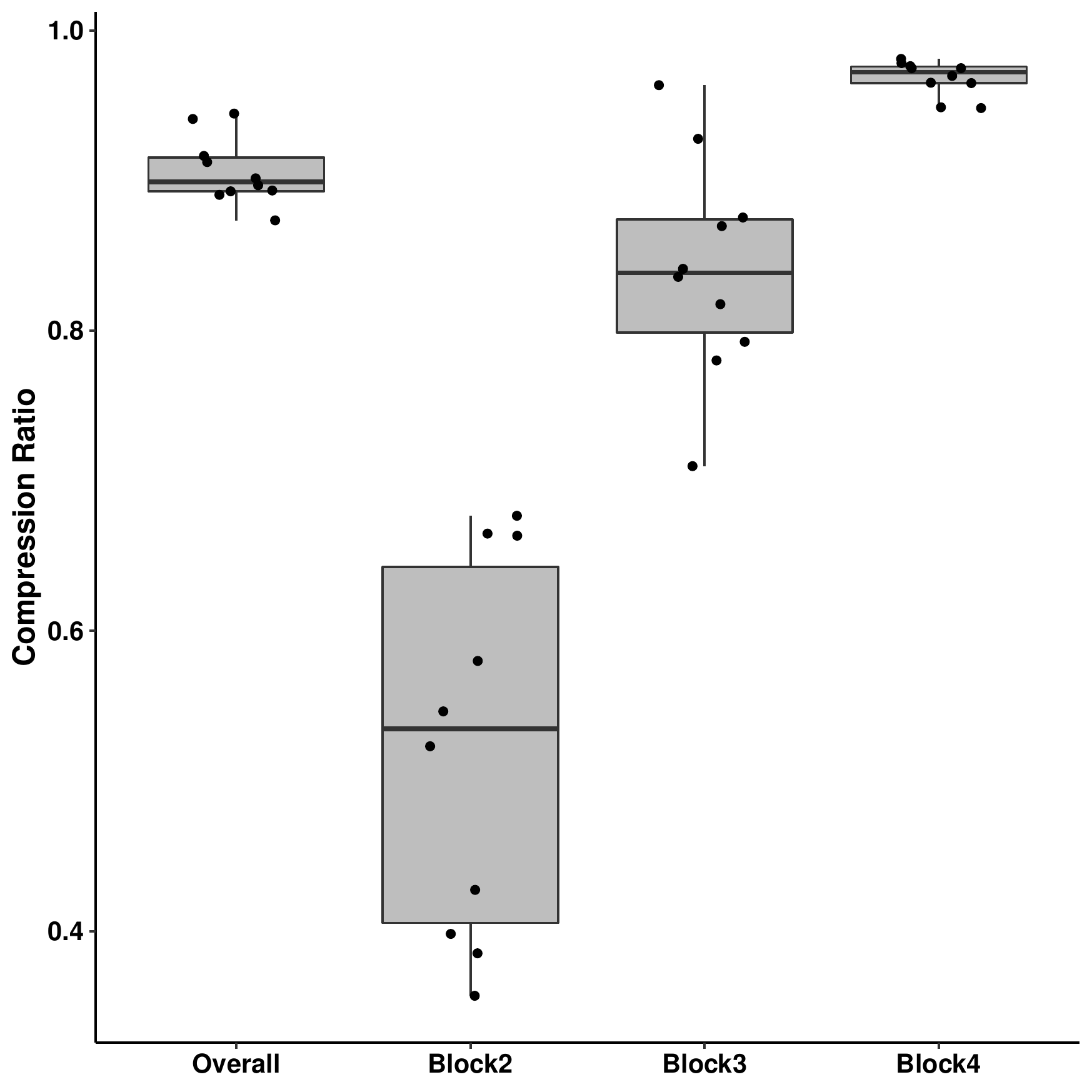}
\label{fig:resnet50-addtree-blockwise-analysis}}
\caption{Comparison results on pruning VGG16 and ResNet50 using different algorithms. For VGG16, there are 49 hyper-parameters in total, 42 of which are continuous parameters lying in $[0,1]$. 
For ReseNet50, there are 63 hyper-parameters, 56 of which are continuous parameters lying in $[0,1]$. In the first column (\Cref{fig:model-pruning-vgg16-cummax-300-iterations,fig:model-pruning-resnet50-cummax-300-iterations}), the solid line is the incumbent median of 10 independent runs, and the shaded region is 95\% confidence interval of median. The second column (\Cref{fig:vgg16-compare-different-algorithms,fig:resnet50-compare-different-algorithms}) shows the best results among all iterations. The third column shows the per-block compression ratio for VGG16 and ResNet50 of \textbf{addtree} using the best parameters. }
\label{fig:pruning-results}
\end{figure*}

We run all algorithms for 300 iterations and the first 50 iterations are random initialization. The results of pruning VGG16 and ResNet50 are shown in \Cref{fig:pruning-results}. We note the public \textbf{smac} implementation cannot handle problems with more than 40 hyper-parameters. To make a comparison possible, we fix some continuous parameters to \textbf{smac}'s default value $0.5$. This is of course a biased choice, and changing these fixed parameters to other values may give a better or worse performance. Choosing different hyper-parameters to freeze could also have a huge impact on the performance. 
From \Cref{fig:model-pruning-vgg16-cummax-300-iterations,fig:model-pruning-resnet50-cummax-300-iterations}, it is clear our proposed Add-Tree performs much better than other competing algorithms.
The distribution of optimized objective values from 10 independent runs are shown in \Cref{fig:vgg16-compare-different-algorithms,fig:resnet50-compare-different-algorithms}, which are zoomed-in versions of \Cref{fig:model-pruning-vgg16-cummax-300-iterations,fig:model-pruning-resnet50-cummax-300-iterations} at the $300$-th iteration, respectively. In \Cref{fig:vgg16-addtree-blockwise-analysis,fig:resnet50-addtree-blockwise-analysis}, we show the blockwise compression analysis of pruning VGG16 and ResNet50 and we see later blocks have much more verbosity than earlier blocks. At the $300$-th iteration, the average running time of different methods is shown in \Cref{tab:model-pruning-average-running-time-300-iteration}. The reason \textbf{random} takes longer time than \textbf{tpe} is the implementation of \textbf{random} comes from SMAC3 and their exist some computational overheads. Regarding the average time of running all algorithms for 300 iterations, \textbf{addtree} is comparable to \textbf{smac} and \textbf{tpe} is comparable to \textbf{random}. Since \textbf{addtree} can achieve a comparable performance at the $100$-th iteration, we also show the average running time of \textbf{addtree} at the $100$-th iteration in parentheses.


\begin{table}[H]
\caption{\label{tab:model-pruning-average-running-time-300-iteration} Average Running Time (s) at the $300$-th Iteration}
\centering
\begin{tabular}{>{\bfseries}lllll}
\toprule
\textbf{Net} & \textbf{addtree} & \textbf{random} & \textbf{smac} & \textbf{tpe} \\
\midrule
VGG16        & 20282 (7391)            & 15889           & 20153         & 15303        \\
ResNet50     & 65494 (20205)           & 58477           & 68824         & 55877        \\
\bottomrule
\end{tabular}
\end{table}

As stated earlier, TPE doesn't model the dependencies between parameters, and effectively performs multiple non-parametric one-dimensional regressions, each regression corresponding to one continuous parameter. 
The results shown here should be compared with the synthetic experiment in \Cref{sec:synthetic-function} and the simple model compression in \Cref{sec:model-compression}. In \Cref{fig:exp:synthetic-optimization-comparison}, \textbf{tpe} performs well among all competing algorithms and this is because parameters of
the synthetic function in \Cref{fig:synthetic-function-jenatton2017} are indeed independent with each other. In \Cref{fig:exp:fc3-compression-comparison}, \textbf{tpe} is also very competitive, possibly because for every layer in FC3, there is only one continuous parameter. 
While in pruning VGG16 and ResNet50, \textbf{tpe} performs much worse than \textbf{addtree}, and this indicates there are complex interactions between parameters in one block.

In \Cref{tab:selected-prune-methods}, we show the optimized discrete choices of pruning methods for each block. M1, M2, M3 indicate the selected pruning methods for the second, the third, and the forth block respectively. From \Cref{tab:selected-prune-methods}, it is clear that using a common compression method across all layers is not an optimal choice.

\begin{table}[H]
\caption{\label{tab:selected-prune-methods}Optimized Pruning Methods at the End of Optimization}
\centering
\begin{tabular}{>{\bfseries}llllrl}
\toprule
\textbf{Algo} & \textbf{M1} & \textbf{M2} & \textbf{M3} & \textbf{Obj} & \textbf{Model}\\
\midrule
\rowcolor{gray!6}  addtree & L1 & L2 & L1 & 1.88 & VGG16\\
addtree & L1 & L1 & L1 & 1.86 & ResNet50\\
\rowcolor{gray!6}  random & L1 & L1 & L1 & 1.78 & VGG16\\
random & L1 & L1 & L1 & 1.77 & ResNet50\\
\rowcolor{gray!6}  smac & L1 & L2 & L2 & 1.81 & VGG16\\
smac & L1 & L1 & L1 & 1.75 & ResNet50\\
\rowcolor{gray!6}  tpe & L1 & L2 & L1 & 1.84 & VGG16\\
tpe & L2 & L1 & L1 & 1.81 & ResNet50\\
\bottomrule
\end{tabular}
\end{table}

\subsection{Neural Architecture Search}
\label{sec:resnet20-nas}

Neural architecture search (NAS) tries to automate the process of designing novel and efficient neural network architectures. In this section, we apply our method to search for activation functions of ResNet20 and their corresponding hyper-parameters on CIFAR-10 dataset. ResNet20 is composed of three main blocks, each of which can further be decomposed into several smaller basic blocks, and every basic block has two ReLU activation functions. An interesting question is whether using other activation functions can improve the performance. 
In this experiment, we limit the activation function to be exponential linear unit (ELU) \cite{clevert2016} or Leaky ReLU (LReLU) \cite{Maas2013RectifierNI}, and all basic blocks in one main block share the activation function and corresponding hyper-parameters. The final search space is parametrized by: (1) discrete choices of activation functions, which could be ELU or LReLU; (2) corresponding hyper-parameters (slope parameter for LReLU, $\alpha$ for ELU), which are 2 continuous parameters constrained in $[0,1]$ in each main block. For this problem, there are 7 categorical parameters and 28 continuous parameters. The tree structure of this problem has a depth of 4, and each node except the root is associated with a 2-dimensional continuous parameter.

We run all algorithms for 100 iterations and the first 50 iterations are random initialization. Every iteration in the hyper-parameter optimization is as follows. First we use the sampled (initialization stage) or optimized (optimization stage) hyper-parameters to construct a ResNet20. Then we train this network on CIFAR-10. For learning, we use stochastic gradient descent with momentum with mini-batches of 128 samples for 10 epochs. The start learning rate is set to be 0.1 and is decreased down to be 0.01 and 0.001 in the beginning of the 7-th epoch and 9-th epoch respectively. The momentum is fixed to be 0.9. Last we test this trained network and use its accuracy on the validation set of CIFAR-10 as our objective.

\begin{figure*}[h]
\centering
\subfloat[ResNet20: all iterations]{
\includegraphics[width=0.31\textwidth]{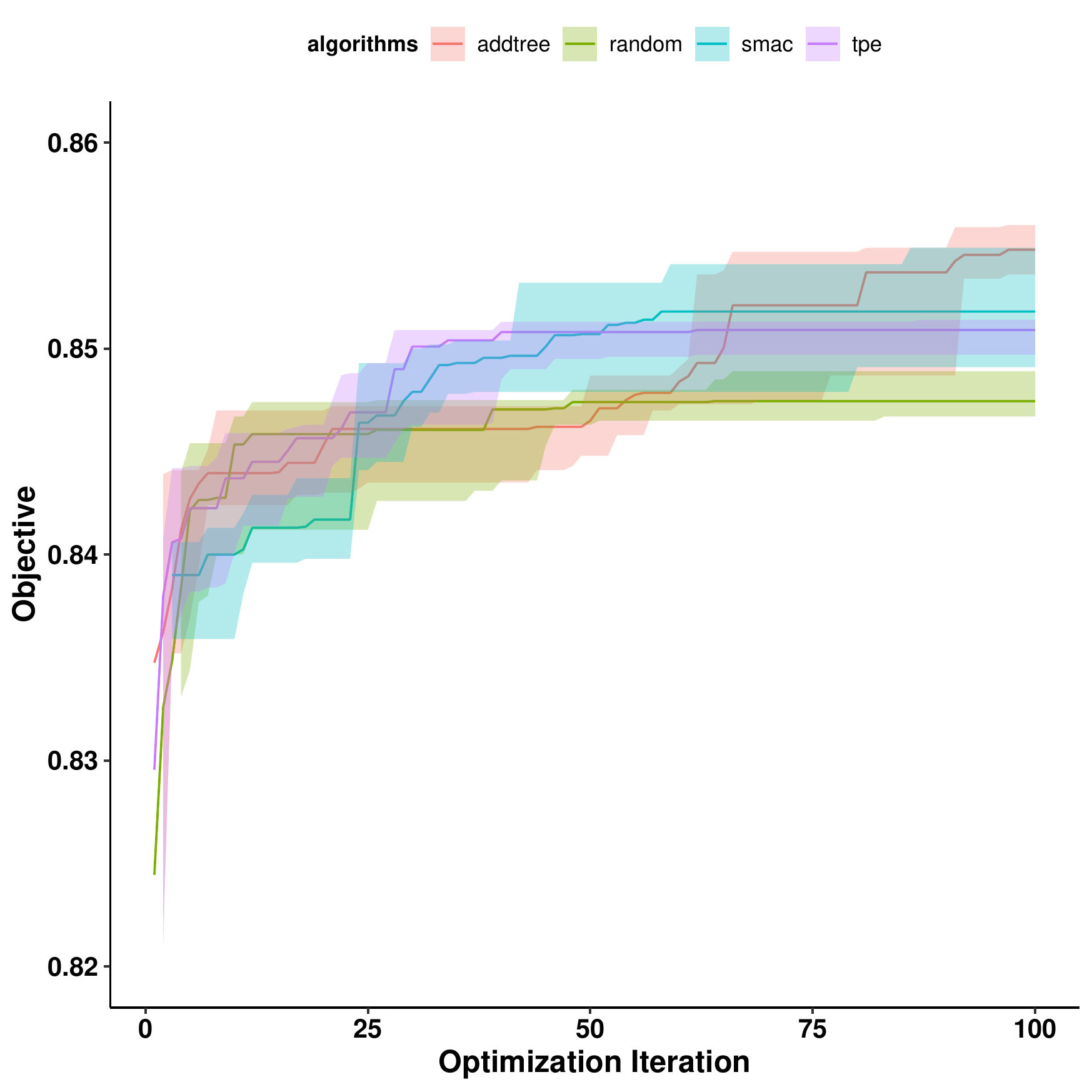}
\label{fig:resnet20-nas-cummax-100-iterations}}
\hfil
\subfloat[ResNet20: best of all iterations]{
\includegraphics[width=0.31\textwidth]{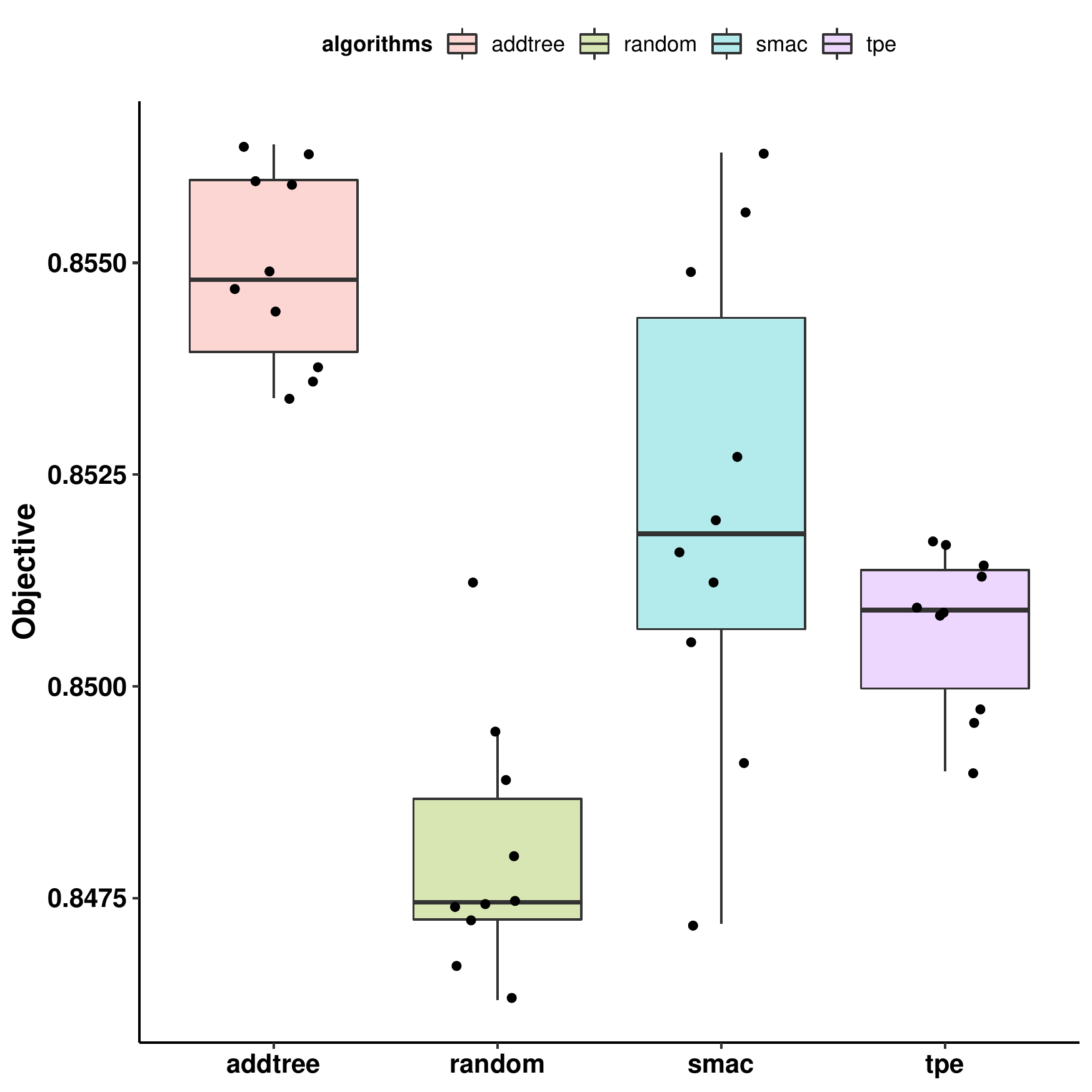}
\label{fig:resnet20-nas-compare-different-algorithms}}
\hfil
\subfloat[ResNet20: running time]{
\includegraphics[width=0.31\textwidth]{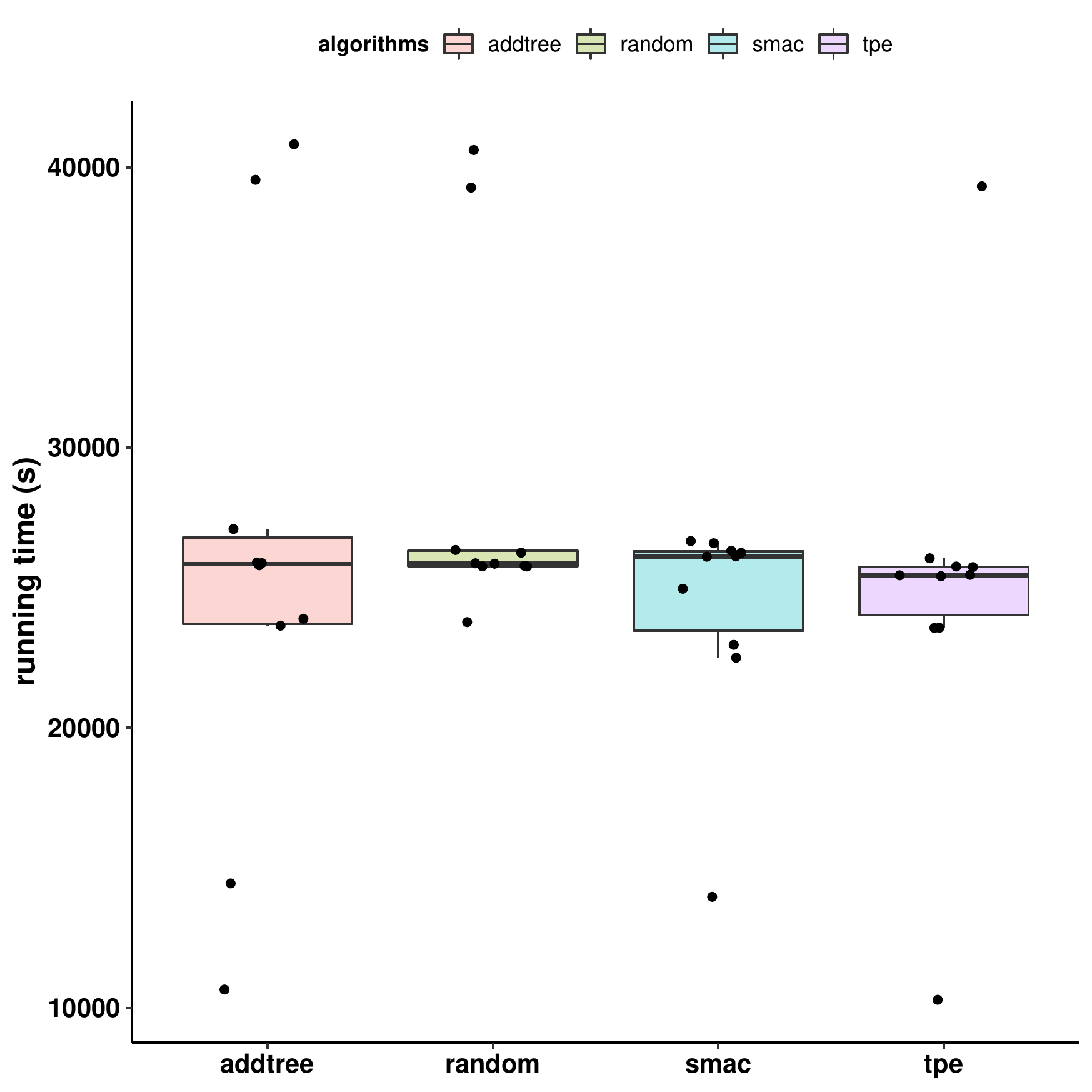}
\label{fig:resnet20-nas-running-time}}
\caption{Comparison results of NAS on ResNet20 using different algorithms. There are 35 hyper-parameters in total, 28 of which are continuous parameters lying in $[0,1]$. In \Cref{fig:resnet20-nas-cummax-100-iterations}, the solid line is the incumbent median of 10 independent runs, and the shaded region is 95\% confidence interval of median. \Cref{fig:resnet20-nas-compare-different-algorithms} shows the best results among all iterations. In \Cref{fig:resnet20-nas-running-time}, we show the running time of different methods.}
\label{fig:resnet20-nas-results}
\end{figure*}

The results of optimizing ResNet20 is shown in \Cref{fig:resnet20-nas-results}. We see our proposed Add-Tree is significantly better than \textbf{tpe} and \textbf{random}. Since there is a overlapping between \textbf{smac} and \textbf{addtree} in \Cref{fig:resnet20-nas-compare-different-algorithms}, we cannot claim our method performs significantly between than \textbf{smac}. 
\Cref{fig:resnet20-nas-running-time} shows the running time of different algorithms at the $100$-th iteration. In this experiment, the running time of all methods is comparable, and this is because we run all algorithms for only 100 iterations and evaluating the expensive objective function dominates the inference time of the random forest, the GP and the Parzen estimator in \textbf{smac}, \textbf{addtree} and \textbf{tpe} respectively.
The existence of several extreme outliers in \Cref{fig:resnet20-nas-running-time} is due to we running these algorithms in a computer cluster managed by a software called HTCondor, which can assign jobs to very fast or very slow machines in a totally random way. 

\begin{table}[H]
\caption{\label{tab:resnet20-nas-optimized-parameters} Optimized Hyper-parameters Using Our Method}
\centering
\begin{tabular}{>{\bfseries}llll}
\toprule
\textbf{Block} & \textbf{Act} & \textbf{\#1 Param} & \textbf{\#2 Param}\\
\midrule
Block 1 & ELU & 0.01 & 0.75\\
Block 2 & LReLU & 0.01 & 0.99 \\
Block 3 & ELU & 0.01 & 0.01\\
\bottomrule
\end{tabular}
\end{table}
\Cref{tab:resnet20-nas-optimized-parameters} shows the optimized hyper-parameters using our algorithm. The first row of \Cref{tab:resnet20-nas-optimized-parameters} means for every basic block in the first main block in ResNet20, ELU is selected, and the hyper-parameters $\alpha$ of the first ELU and the second ELU are 0.01 and 0.7484 respectively.
During the optimization, we train ResNet20 for only 10 epochs in order to speed up this process, and the accuracy shown in \Cref{fig:resnet20-nas-cummax-100-iterations} underestimate the true accuracy. Thus we construct a new ResNet20 based on the optimized hyper-parameters shown in \Cref{tab:resnet20-nas-optimized-parameters}, and train this network for 200 epochs. We decrease the learning to 0.1 and 0.001 in the beginning of the 100-th epoch and 150-th epoch respectively and also add a L2-weight decay regularization term, which is set to be 0.0001. The final test error is 7.74\%, compared with the original test error (8.27\%), our optimized ResNet20 performs much better. 

\section{CONCLUSION}
\label{sec:conclusion}

In this work, we have designed a covariance function that can explicitly utilize the problem structure, and demonstrated its efficiency on a range of problems. In the low data regime, our proposed Add-Tree covariance function enables an explicit information sharing mechanism, which in turn makes BO more sample-efficient compared with other model based optimization methods. 
Contrary to other GP-based BO methods, we do not impose restrictions on the structure of a conditional parameter space, greatly increasing the applicability of our method. 
We also directly model the dependencies between different observations under the framework of Gaussian Processes, instead of placing parametric relationships between different paths, making our method more flexible. In addition, we incorporate this structure information and develop a parallel algorithm to optimize the acquisition function. 
For both components of BO, our proposed method allows us to work in a lower dimensional space compared with the dimension of the original parameter space.

Empirical results on an optimization benchmark function, a simple neural network compression problem, on pruning VGG16 and ResNet50, and on searching for activation functions of ResNet20 show our method significantly outperforms other competing model based optimization algorithms in conditional parameter spaces, including SMAC, TPE and~\cite{jenatton2017}.

\ifCLASSOPTIONcompsoc
  \section*{Acknowledgments}
\else
  \section*{Acknowledgment}
\fi

Xingchen Ma is supported by Onfido.
This research received funding from the Flemish Government under the “Onderzoeksprogramma Artifici\"{e}le Intelligentie (AI) Vlaanderen” programme.

\ifCLASSOPTIONcaptionsoff
  \newpage
\fi



\bibliographystyle{IEEEtran}
\bibliography{IEEEabrv,mybib.bib}

\begin{IEEEbiographynophoto}{Xingchen Ma}
is a PhD student in the Electrical Engineering department at KU Leuven.
\end{IEEEbiographynophoto}

\begin{IEEEbiographynophoto}{Matthew Blaschko}
is a faculty member in the Electrical Engineering department at KU Leuven.
\end{IEEEbiographynophoto}


%

\newpage
\clearpage

\appendices

\section{Proofs of Theoretical Results}
To prove \Cref{prop:add-tree-is-psd}, we make use of: (1) the sum of two valid kernels is a valid kernel; (2) the product of two valid kernels is a valid kernel\cite{rasmussen2006}. 
 
\begin{IEEEproof}[Proof of \Cref{prop:add-tree-is-psd}]
We will consider each term in \Cref{eq:add-tree-kernel-1} and demonstrate that it results in a positive semi-definite covariance function over the whole set of data points, not just the data points that follow the given path.  In particular, consider the p.s.d.\ covariance function $k_{v}^\delta(\bm{x}_{i'},\bm{x}_{j'}) = \begin{cases} 1 & \text{ if } v\in l_i \wedge v \in l_j \\
0 & \text{ otherwise}
\end{cases}$, for some vertex $v$.  We see that the product $k_v^c \times k_{v}^\delta$ defines a p.s.d.\ covariance function over the entire space of observations (since the product of two p.s.d.\ covariance functions is itself p.s.d.). 
In this way, we may interpret \Cref{eq:add-tree-kernel-1} as a summation over only the non-zero terms of a set of covariance functions defined over all vertices in the tree. As the resulting covariance function sums over p.s.d.\ covariance functions, and positive semi-definiteness is closed over positively weighted sums, the result is p.s.d.
\end{IEEEproof}

In order to prove a tree-structured function $f_{\mathcal{T}}$ has bounded RKHS norm in an RKHS with an Add-Tree covariance function $k_{\mathcal{T}}$, we need the following \Cref{prop:wainwright2019-prop}.

\begin{proposition}[Proposition 12.27 in \cite{wainwright2019}]
\label{prop:wainwright2019-prop}
Suppose that $\mathcal{H}_1$ and $\mathcal{H}_2$ are both RKHSs with kernels $k_1$ and $k_2$ respectively.
Then the space $\mathcal{H} =\mathcal{H}_1  + \mathcal{H}_2$ with norm:
\begin{equation}
    \norm{f}_{\mathcal{H}}^2 \coloneqq \min_{\substack{f=f_1+f_2 \\f_1 \in \mathcal{H}_1, f_2 \in \mathcal{H}_2  }}   \{ \norm{f_1}_{\mathcal{H}_1}^2 + \norm{f_2}_{\mathcal{H}_2}^2 \}
\end{equation}
is an RKHS with kernel $k = k_1 + k_2$.
\end{proposition}

Equipped with this proposition, we can now prove \Cref{prop:RKHS-bound-addtree-function}.

\begin{IEEEproof}[Proof of \Cref{prop:RKHS-bound-addtree-function}]
Taking an arbitrary path $l_i$, for $j \in \{1,\dots, h_i\}$, where $h_i$ is the length of path $l_i$ and $h_i \le M$.
From \Cref{prop:wainwright2019-prop}, we have $f_i \in \mathcal{H}_i$, where $\mathcal{H}_i$ is a RKHS with kernel $k_i$.
Due to the additive assumption in \Cref{eq:additive-assumption}, $H_{ij}$ and $H_{ik}$ share only zero function by construction, that is $\mathcal{H}_{ij} \cap \mathcal{H}_{i,k} = \{ 0 \}$, for all $j \neq k$, we have:
\begin{equation*}
    \norm{f_{i}}_{k_i}^2 = \sum_{j=1}^{h_i} \norm{f_{ij}}_{k_{ij}}^2 \leq h_i \tilde{B} \leq M \tilde{B}.
\end{equation*}
By \Cref{def:TreeStructuredFunction},  $\norm{f_\mathcal{T}}_{k_{\mathcal{T}}}^2 \le M \tilde{B}$. 
\end{IEEEproof}

Since the regret bound requires the bound of the tail sum of operator spectrum of $k_\mathcal{T}$, we will need the following lemma.

\begin{lemma}[Weyl’s inequality\cite{fulton2000}]
\label{lemma:weyl-inequality}
If $A_1$, $A_2$, and $A_3$ are Hermitian $n$ by $n$ matrices with $A_3=A_1+A_2$, we denote the eigenvalues of $A_1$, $A_2$, $A_3$ by $\{ \alpha_1^{(1)},\dots,\alpha_n^{(1)} \}$, $\{ \alpha_1^{(2)},\dots,\alpha_n^{(2)} \}$,$\{ \alpha_1^{(3)},\dots,\alpha_n^{(3)} \}$ respectively, and eigenvalues of $A_1$, $A_2$, $A_3$ are listed in decreasing order: $\alpha_1^{(1)} \ge \dots \ge \alpha_n^{(1)}, \alpha_1^{(2)} \ge \dots \ge \alpha_n^{(2)}, \alpha_1^{(3)} \ge \dots \ge \alpha_n^{(3)}$, then $\alpha_{i+j-1}^{(3)} \le \alpha_i^{(1)} + \alpha_j^{(2)}$, whenever $i+j-1 \le n$.
\end{lemma}

To generalize \Cref{lemma:weyl-inequality} to the Gram matrix of an Add-Tree covariance function $k_{\mathcal{T}}$, firstly we use a different notation to list the covariance functions defined on the vertices of $\mathcal{T}$. 
Let $k_{\mathcal{T}}^{(h,j)}$ be a covariance function defined on the continuous variables at the $j$-th vertex in the $h$-th level of $\mathcal{T}$, $n_h$ be the number of vertices in the $h$-th level of $\mathcal{T}$, $\bm{K}_{\mathcal{T}}$ be the Gram matrix of $k_{\mathcal{T}}$, $\{\lambda_1,\dots,\lambda_n\}$ be the eigenvalues of $\bm{K}_{\mathcal{T}}$ with $\lambda_1 \ge \dots \ge \lambda_n$, $\bm{K}_{\mathcal{T}}^{(h,j)}$ be the Gram matrix of $k_{\mathcal{T}}^{(h,j)}$, $\{\lambda_1^{(h,j)},\dots,\lambda_{n_{(h,j)}}^{(h,j)}\}$ be the eigenvalues of $\bm{K}_{\mathcal{T}}^{(h,j)}$ with $\lambda_1^{(h,j)} \ge \dots \ge \lambda_{n_{(h,j)}}^{(h,j)}$, where $\sum_{j=1}^{n_h} {n_{(h,j)}} = n$ for all $h \in \{ 1,\dots,M \}$ by the additive assumption in \Cref{eq:additive-assumption} and the construction of $k_{\mathcal{T}}$ in \Cref{def:TreeStructuredFunction}.
The collection of $k_{\mathcal{T}}^{(h,j)}$ is a regular set, instead of a multiset, because different levels of $\mathcal{T}$ cannot share variables. 

\begin{proposition}
\label{prop:weyl-extension}
For all $k_i \in \mathbb{N}$ such that $1 + \sum_{i=1}^{M} k_i \le n$, we have:
\begin{equation}
    \lambda_{1 + \sum_{i=1}^{M} k_i} \le \sum_{h=1}^{M} \lambda_{1 + k_i}^{(h)},
\end{equation}
where $\{ \lambda_1^{(h)}, \dots, \lambda_{n}^{(h)} \}$ is the union of eigenvalues of all covariance functions in the $h$-th level of $\mathcal{T}$ with $\lambda_1^{(h)} \ge \dots \ge \lambda_n^{(h)}$.
\end{proposition}

To illustrate \Cref{prop:weyl-extension}, recall in \Cref{eq:joint-distributation}, the Gram matrix of $k_{\mathcal{T}}$ is decomposed into two parts:
\begin{align*}
    \bm{K}_{\mathcal{T}} &=\begin{bmatrix} 
    \bm{K}_r^{(11)} & \bm{K}_r^{(12)} \\ 
    \bm{K}_r^{(21)} & \bm{K}_r^{(22)} 
    \end{bmatrix} + \begin{bmatrix}
           \bm{K}_1  & \bm{0}  \\
            \bm{0}    & \bm{K}_2 
         \end{bmatrix} \\
         &= \bm{K}_{\mathcal{T}}^{(1,1)} + \begin{bmatrix}
           \bm{K}_{\mathcal{T}}^{(2,1)}  & \bm{0}  \\
            \bm{0}    & \bm{K}_{\mathcal{T}}^{(2,2)}
         \end{bmatrix}.
\end{align*}
The union of eigenvalues of covariance functions in the second level of $\mathcal{T}$ is the union of eigenvalues of $\bm{K}_{\mathcal{T}}^{(2,1)}$ and $\bm{K}_{\mathcal{T}}^{(2,2)}$.
In general, the deeper the level in $\mathcal{T}$, the sparser the decomposed Gram matrix.

Equipped with the above propositions, we can now prove our main theoretical result.

\begin{IEEEproof}[Proof of \Cref{th:bounds-kernels}]
Explicit bounds on the eigenvalues of the squared exponential kernel w.r.t the Gaussian covariate distribution $\mathcal{N}(0, (4a)^{-1} I_d)$ have been given in \cite{seeger2008}. For a $d$ dimensional squared exponential kernel, we have:
\begin{equation*}
    \lambda_{s} \leq c^d B^{s^{1/d}},
\end{equation*}
where $c,B$ are only dependent on $a,d$ and kernel's lengthscales $\theta$. 
As pointed out in \cite{srinivas2010}, the same decay rate of $\lambda_s$ holds w.r.t uniform covariate distribution.
From \Cref{th:srinivas2010-8}, to bound $\gamma_T$, we only need to bound $B_{\kT}(T_*)$, the tail sum of the operator spectrum of $\kT$. By \Cref{prop:weyl-extension}, we have\footnote{The standard numerical method shows $\frac{1}{n} \lambda^{\text{mat}}_s$ (the $s$-th eigenvalue of the Gram matrix) will converge to $\lambda_s$ (the $s$-th eigenvalue of the corresponding kernel) in the limit that the number of observations goes to infinity\cite{rasmussen2006}. It can be expected that $\lambda_s < \lambda^{\text{mat}}_s$ given a large number of observations.}, 
\begin{equation}
B_{\kT}(T_*) = \sum_{s>T_*} \lambda_s \le \sum_{s>T_*} \sum_{h=1}^M \lambda_{\floor*{s/M}}^{(h)}.
\end{equation}

Let $s_+ = \floor*{s/M}, T_+ = \floor*{T_* / M}$, then 
\begin{align}
\label{eq:main-rbf-step1}
B_{\kT}(T_*) &\leq M \sum_{s_+ > T_+} \sum_{h=1}^M \lambda_{s_+}^{(h)} \nonumber \\
         &\leq M \sum_{s_+ > T_+} \sum_{h=1}^M c^{d} B^{s_{+}^{1/d}}  \nonumber \\
         &\leq M^2 c^{\td} \int_{T_+}^{\infty} \exp (s_{+}^{1/\td} \log B) ds_{+}.
\end{align}

To simplify the above integral, let $\alpha = -\log B$ and substitute $t = \alpha s_{+}^{1/\td}$, then $ds_{+} = \frac{\td}{\alpha} (\frac{t}{\alpha})^{\td-1} dt$, we have:
\begin{align}
B_{\kT}(T_*) &\leq \frac{M^2 c^{\td} \td}{\alpha^{\td}} \int_{\alpha T_{+}^{1/\td}}^{\infty} e^{-t} t^{\td-1} dt \nonumber \\
         &= \frac{M^2 c^{\td} \td}{\alpha^{\td}} \Gamma(\td, \alpha T_{+}^{1/\td}).
\end{align}
where $\Gamma(d, \beta) = \int_{\beta}^{\infty} e^{-t} t^{d-1} dt$ is the incomplete Gamma function \cite{seeger2008} and using $\Gamma(d, \beta)  = (d-1)! e^{-\beta} \sum_{k=0}^{d-1} \beta^{k} / k!$, let $\beta = \alpha T_{+}^{1/\td}$, we have:
\begin{align}
\label{eq:bound-of-BT}
B_{\kT}(T_*) &\leq  \frac{M^2 c^{\td} \td!}{\alpha^{\td}} e^{-\beta} \sum_{k=0}^{\td-1} \frac{\beta^{k}}{k!} 
\end{align}

Following \cite{berkenkamp2019}, we have
\begin{align}
    \frac{1}{\alpha^{\td}} &= \frac{1}{(-\log B)^{\td}} = \mathcal{O}(g(t)^{2 \td}),  \label{eq:bound-of-aux-1}\\
    c^{\td} &= (\frac{2a}{A})^{\td/2} = \mathcal{O}(g(t)^{-\td}). \label{eq:bound-of-aux-2}
\end{align}
Plugging \Cref{eq:bound-of-aux-1,eq:bound-of-aux-2} into \Cref{eq:bound-of-BT}, we have:
\begin{equation*}
    B_{\kT}(T_*) = \mathcal{O}(\td! M^2 g(t)^{\td}  e^{-\beta} \beta^{\td-1}).
\end{equation*}
Following \cite{srinivas2010} and \cite{berkenkamp2019}, we use $T_+ = [ \log (T n_T) / \alpha ]^{\td}$, so that $\beta = \log (T n_T)$ and doesn't depend on $g(t)$. Following \cite{srinivas2010}, it turns out $\tau=d$ is the optimal choice, and we can ignore the second part of the right hand side of \Cref{eq:srinivas2010-max-gain}. 
To obtain the bound of $\gamma_t$, we decompose the first part of the right hand side of \Cref{eq:srinivas2010-max-gain} into two parts:
\begin{align*}
    S_1 &= T_{*} \log (\frac{r n_T}{\sigma^2}), \\
    S_2 &= C_4 \sigma^2 (1-r/T)( B_{\kT}(T_*) T^{\tau + 1} + 1 ) \log T.
\end{align*}
In the following, we bound $S_1$ and $S_2$ respectively. 
\begin{align}
    S_1 &\leq (M+1) T_{+} \log (r n_T) \nonumber \\
        &= \mathcal{O} \left( M (\log (T n_T))^{\td}  g(t)^{2\td} \log (r n_T) \right) \nonumber \\
        &= \mathcal{O} \left( M ((\td+1) \log T )^{\td} g(t)^{2\td} (\log r + \td \log T)    \right) \nonumber \\
        &= \mathcal{O} \bigg( M \td^{\td+1} (\log T)^{\td+1} g(t)^{2\td} \nonumber \\
        &\qquad \quad +  M \td^{\td} (\log T)^{\td} g(t)^{2\td} \log r \bigg). \label{eq:bound-S1}
\end{align}
Plugging the bound for $B_{\kT}(T_*)$ into $S_2$, we obtain the bound for $S_2$:
\begin{align}
    S_2 =& \mathcal{O} \bigg( (1 - \frac{r}{T}) ( \td! M^2 T^{\td+1} g(t)^{\td} \frac{1}{T^{\td+1} \log T} \nonumber  \\
    &\qquad \times \td^{\td} (\log T)^{\td-1}  \log T ) \bigg) \nonumber \\
    &= \mathcal{O} \left(  (1 - \frac{r}{T}) \td! M^2 g(t)^{\td} \td^{\td} (\log T)^{\td-1}  \right) \label{eq:bound-S2}
\end{align}
Since $g(t)>1$ and is an increasing function, $S_1$ dominates $S_2$, let $r=T$, we obtain the bound for $\gamma_T$:
\begin{align*}
\gamma_T &= \mathcal{O} \left( M \td^{\td + 1} (\log T)^{\td + 1} g(t)^{2 \td} \right) \\
        &= \mathcal{O} \left( d \td^{\td} (\log T)^{\td+1} g(t)^{2 \td} \right).
\end{align*}

To obtain the regret for Mat\'{e}rn case, we follow the proof in \cite{seeger2008} and \cite{berkenkamp2019}, and have bounds on the eigenvalues of the Mat\'{e}rn covariance function w.r.t. bounded support measure $\mu_T$, 
\begin{equation*}
    \lambda_{s | \mu_T} \le C(1+\delta) s^{-(2 \nu + d)/d} \quad \forall s>s_0.
\end{equation*}

Similar to \Cref{eq:main-rbf-step1} in \Cref{th:bounds-kernels}, we have:
\begin{equation}
    B_{k_{\mathcal{T}}}(T_*) \le M^2 B_k(T_+).
\end{equation}
where $B_k(T_+)$ is derived in \cite{berkenkamp2019},
\begin{equation*}
    B_k(T_+) = \mathcal{O}(g(t)^{2 \nu + \td} T_{+}^{1 - (2 \nu + \td) / \td}).
\end{equation*}
As with \cite{srinivas2010}, we choose $\tau=2 \nu \td /(2 \nu + \td (1+\td))$, $T_+ = (T n_T )^{\eta} (\log(T n_T))^{-\eta} $, where $\eta=\td / (2 \nu + \td)$, and obtain the bound for $\gamma_T$:
\begin{align}
    \max_{r=1,\dots,T} \bigg( &M T_+ \log(r n_T) + (1-r/T) \nonumber \\ 
    &\times (M^2 g^{2\nu+\td} T_{+}^{1-(2\nu+\td)/\td} T^{\tau+1} + 1) \log(T) \bigg) \nonumber \\
    &+ \mathcal{O}(T^{1-\tau/\td}) 
    \label{eq:matern-bound-1}
\end{align}

Since the inner part of \Cref{eq:matern-bound-1} is a concave function w.r.t $r$, we can obtain its maximum easily. 
By setting the derivative of the inner part \Cref{eq:matern-bound-1} w.r.t $r$ to be zero, we have 
\begin{align*}
    \gamma_T = \mathcal{O}(M^2 g(t)^{2 \nu+\td} (\log T) T^{\frac{\td(\td+1)}{2 \nu + \td(\td+1)}}).
\end{align*}

\end{IEEEproof}

\Cref{coro:cumulative-regret} follows from \Cref{th:bounds-kernels} and \Cref{th:berkenkamp2019-1}.

\begin{IEEEproof}[Proof of \Cref{coro:cumulative-regret}]
The first part of \Cref{eq:th-berkenkamp2019-1} can be bounded using a constant and we only need to bound the second term $\sqrt{t \beta_t I_{\theta_t} (\bm{y}_t,f)}$. 
From \Cref{eq:maximum-mutual-definition}, $I_{\theta_t} (\bm{y}_t,f)$ can be bounded by the maximum information gain $\gamma_t$, which has been given in \Cref{th:bounds-kernels}.
Combine with \Cref{th:berkenkamp2019-1}, using the RKHS bound of a tree-structured function in \Cref{prop:RKHS-bound-addtree-function}, we obtain bounds of the cumulative regret of our proposed Add-Tree covariance function when each covariance function defined on vertices in $\mathcal{T}$ is a squared exponential kernel or a Mat\'{e}rn kernel.
\end{IEEEproof}

\section{Implementation Details}
\label{sec:implementation-details}

The most important part in our implementation is a linear representation for the tree structure and observations in a tree-structured parameter space using breadth-first search (BFS). When BFS encounters a node, the linearization routine adds a tag, which is the rank of this node in its siblings, in front of the variable associated with this node, and this tag will be used to construct the delta kernel in \Cref{algo:kernel-construction}. \Cref{fig:bfs-example} shows the liner representation corresponding to the tree structure in \Cref{fig:example}. 

\begin{figure}[h]
\centering
\resizebox{0.95\columnwidth}{!}{
\begin{tikzpicture}

\fill[blue!40!white] (0,0) rectangle (1,1);
\draw (0,0) rectangle (1,1) node [pos=.5] {0};
\draw (1,0) rectangle (3,1) node [pos=.5] {$\bm{v}_r$};

\fill[blue!40!white] (3,0) rectangle (4,1);
\draw (3,0) rectangle (4,1) node [pos=.5] {0};
\draw (4,0) rectangle (6,1) node [pos=.5] {$\bm{v}_{p_1}$};

\fill[blue!40!white] (6,0) rectangle (7,1) ;
\draw (6,0) rectangle (7,1) node [pos=.5] {1};
\draw (7,0) rectangle (10,1) node [pos=.5] {$\bm{v}_{p_2}$};

\end{tikzpicture}
}
\caption{Linear representation of the tree structure corresponding to \Cref{fig:example}} 
\label{fig:bfs-example}
\end{figure}

When linearizing an observation, we modify the tag in \Cref{fig:bfs-example} using the following rules:
\begin{itemize}
    \item If a node is not associated with a continuous parameter, its tag is changed to a unique value.
    \item If a node doesn't in this observation's corresponding path, its tag is changed to a unique value.
    \item Otherwise, we set the values after this tag to be the sub-parameter restricted to this node.
\end{itemize}{}

For example, the linear representation of an observation $(0.1,0.2,0.3,0.4)$ falling into the left path is shown in \Cref{fig:bfs-example-left-observation} and the linear representation of an observation $(0.5,0.6,0.7,0.8,0.9)$ falling into the right path is shown in \Cref{fig:bfs-example-right-observation}.

\begin{figure}[h]
\centering
\resizebox{0.95\columnwidth}{!}{
\begin{tikzpicture}

\fill[blue!40!white] (0,0) rectangle (1,1);
\draw (0,0) rectangle (1,1) node [pos=.5] {0};
\draw (1,0) rectangle (2,1) node [pos=.5] {0.1};
\draw (2,0) rectangle (3,1) node [pos=.5] {0.2};

\fill[blue!40!white] (3,0) rectangle (4,1);
\draw (3,0) rectangle (4,1) node [pos=.5] {0};
\draw (4,0) rectangle (5,1) node [pos=.5] {0.3};
\draw (5,0) rectangle (6,1) node [pos=.5] {0.4};

\fill[blue!40!white] (6,0) rectangle (7,1) ;
\draw (6,0) rectangle (7,1) node [pos=.5] {Uniq};
\draw (7,0) rectangle (10,1) node [pos=.5] {$\bm{v}_{p_2}$};

\end{tikzpicture}
}
\caption{Linear representation of an observation $(0.1,0.2,0.3,0.4)$ falling into the lower path corresponding to \Cref{fig:example}} 
\label{fig:bfs-example-left-observation}
\end{figure}

\begin{figure}[h]
\centering
\resizebox{0.95\columnwidth}{!}{
\begin{tikzpicture}

\fill[blue!40!white] (0,0) rectangle (1,1);
\draw (0,0) rectangle (1,1) node [pos=.5] {0};
\draw (1,0) rectangle (2,1) node [pos=.5] {0.5};
\draw (2,0) rectangle (3,1) node [pos=.5] {0.6};

\fill[blue!40!white] (3,0) rectangle (4,1);
\draw (3,0) rectangle (4,1) node [pos=.5] {Uniq};
\draw (4,0) rectangle (6,1) node [pos=.5] {$\bm{v}_{p_1}$};

\fill[blue!40!white] (6,0) rectangle (7,1) ;
\draw (6,0) rectangle (7,1) node [pos=.5] {1};
\draw (7,0) rectangle (8,1) node [pos=.5] {0.7};
\draw (8,0) rectangle (9,1) node [pos=.5] {0.8};
\draw (9,0) rectangle (10,1) node [pos=.5] {0.9};

\end{tikzpicture}
}

\caption{Linear representation of an observation $(0.5,0.6,0.7,0.8,0.9)$ falling into the upper path corresponding to \Cref{fig:example}} 
\label{fig:bfs-example-right-observation}
\end{figure}

Based on this linear representation, it is now straightforward to compute the covariance function, which is constructed in \Cref{algo:kernel-construction},  between any two observations. We note the dimension of such a linear representation has order $\mathcal{O}(d)$, where $d$ is the dimension of the original parameter space, thus there is little overhead compared with other covariance functions in existing GP libraries.

\end{document}